\documentclass[preprint,12pt]{elsarticle}

\usepackage{amsmath}
\usepackage{amssymb}
\usepackage{amsthm}
\usepackage{latexsym} 

\usepackage{graphicx}      
\usepackage{booktabs}      
\usepackage{multirow}      
\usepackage{diagbox}       
\usepackage{adjustbox}     
\usepackage{array}         
\usepackage{makecell}      
\usepackage{stfloats}      
\usepackage{framed}        

\usepackage{caption} 
\usepackage[font=footnotesize,labelformat=simple]{subcaption}

\usepackage{url}           
\usepackage{xcolor}        
\usepackage{bbding}        
\usepackage{geometry}      

\definecolor{Gray}{gray}{0.9}
\definecolor{newcolor}{rgb}{.8,.349,.1}

\usepackage{hyperref}
\usepackage{tabularx}

\journal{Medical Image Analysis}

\begin{document}

\begin{frontmatter}




\title{MICCAI STSR 2025 Challenge: Semi-Supervised Teeth and Pulp Segmentation and CBCT-IOS Registration}


\cortext[cor1]{Corresponding authors: ljun77@hdu.edu.cn (Jun Liu); shuaiwang.tai@gmail.com (Shuai Wang), hz143@leicester.ac.uk (Huiyu Zhou)}

\fntext[eq]{These authors contributed equally to this work.}

\author[1]{Yaqi {Wang}\fnref{eq}}
\author[4]{Zhi {Li}\fnref{eq}} 
\author[12]{Chengyu {Wu}}
\author[1]{Jun {Liu}\corref{cor1}}
\author[3]{Yifan {Zhang}}

\author[5]{Jialuo {Chen}}
\author[5]{Jiaxue {Ni}}
\author[5]{Qian {Luo}}

\author[5]{Jin {Liu}}
\author[18]{Can {Han}}

\author[21]{Changkai {Ji}}
\author[22]{Zhi Qin {Tan}}
\author[23]{Ajo Babu {George}}
\author[24]{Liangyu {Chen}}

\author[11]{Qianni {Zhang}}
\author[18]{Dahong {Qian}}
\author[4]{Shuai {Wang}\corref{cor1}}
\author[110]{Huiyu {Zhou}\corref{cor1}}


\address[1]{Innovation Center for Electronic Design Automation Technology, Hangzhou Dianzi University, Hangzhou, China}
\address[4]{School of Cyberspace, Hangzhou Dianzi University, Hangzhou, China}
\address[5]{Hangzhou Dianzi University, Hangzhou, China}

\address[3]{Hangzhou Geriatric Stomatology Hospital, Hangzhou Dental Hospital Group, Hangzhou, China}
\address[10]{Shenzhen University, Shenzhen, China}
\address[11]{Queen Mary University of London, London, United Kingdom}
\address[12]{Shandong University, Weihai, China}

\address[21]{School of Automation and Intelligent Sensing, Shanghai Jiao Tong University, Shanghai, China}
\address[22]{King’s College London, United Kingdom}
\address[23]{DiceMed, Odisha, India}
\address[24]{College of Information Engineering, China Jiliang University, Hangzhou, China}

\address[110]{University of Leicester, Leicester, United Kingdom}

\begin{abstract}
Cone-Beam Computed Tomography (CBCT) and Intraoral Scanning (IOS) are essential modalities for digital dentistry, yet the scarcity of annotated data hinders the development of automated solutions for complex tasks like pulp canal segmentation and cross-modal registration. To benchmark and advance semi-supervised learning (SSL) in this domain, we organized the Semi-supervised Teeth Segmentation and Registration (STSR 2025) Challenge at MICCAI 2025.
The challenge introduced two distinct tasks: (1) Semi-supervised segmentation of teeth and root pulp canals in CBCT, and (2) Semi-supervised rigid registration of CBCT and IOS data. We provided a dataset comprising 60 labeled and 640 unlabeled IOS samples, alongside 30 labeled and 250 unlabeled CBCT scans with varying resolutions and fields of view. The challenge attracted significant community interest, with top teams submitting open-source solutions that were rigorously evaluated for algorithmic excellence.
All successful submissions leveraged deep learning-based SSL methods. For the segmentation task, participants predominantly employed nnU-Net and State Space Model (SSM) architectures like Mamba, utilizing pseudo-labeling and consistency regularization to effectively exploit unlabeled data. The top segmentation method achieved a Dice score of 0.967 and an Instance Affinity (IA) of 0.738 on the hidden validation set. For the registration task, effective solutions utilized point cloud-based networks such as PointNetLK, combined with differentiable SVD heads and geometric data augmentation to handle large modality discrepancies. The winning registration frameworks demonstrated that hybrid neural-classical refinement strategies could achieve competitive alignment accuracy even with limited ground truth.
Both the challenge dataset and the participants’ submitted code have been made publicly available on GitHub (\url{https://github.com/ricoleehduu/STS-Challenge-2025}), ensuring transparency and reproducibility.
\end{abstract}


\end{frontmatter}



\section{Introduction}
Digital dentistry is undergoing a paradigm shift, driven by the integration of advanced imaging modalities and Artificial Intelligence (AI). While the analysis of external tooth surfaces is well-established, modern clinical workflows increasingly demand a comprehensive understanding of internal anatomical structures and the fusion of multi-modal data~\cite{Schulze2016_CBCT_IOS_registration}. Specifically, the precise delineation of the root canal system is paramount for successful endodontics, determining the efficacy of cleaning and shaping procedures~\cite{Wang2023_RootCanalSegmentation_DentalNet_PulpNet}. Simultaneously, the fusion of Cone-Beam Computed Tomography (CBCT) with Intraoral Scans (IOS) is indispensable for orthodontics and prosthodontics. CBCT provides volumetric data of tooth roots and bone, whereas IOS captures high-resolution surface details of the crown. Integrating these modalities creates a ``digital patient'' model, which is essential for designing accurate implant guides and orthodontic appliances~\cite{Elgarba2024_AI_CBTC_IOS_registration}.
\begin{figure*}[!ht]
\centering
\includegraphics[width=\linewidth]{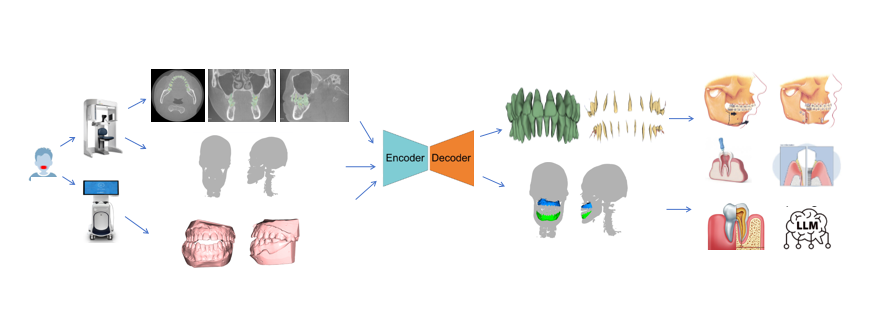}
\caption{Overview of the Semi-Supervised Teeth Segmentation and Registration (STSR 2025) framework and its clinical utility. The workflow proceeds from (Left) multi-modal data acquisition, capturing volumetric Cone-Beam Computed Tomography (CBCT) and surface-based Intraoral Scans (IOS). These inputs are processed in (Middle) the core semi-supervised learning paradigm, utilizing Encoder-Decoder architectures (e.g., U-Mamba, PointNet++) to perform two key tasks: high-precision segmentation of teeth and root pulp canals, and cross-modal rigid registration of IOS crowns with CBCT roots. The resulting 3D models facilitate (Right) diverse downstream clinical applications, including orthodontic treatment planning, endodontic (root canal) therapy, implant surgery simulation, and Large Language Model (LLM)-driven automated diagnostic reporting.}
\label{fig:fig1}
\end{figure*}
However, automating these tasks via deep learning faces significant hurdles, primarily driven by anatomical complexity and the scarcity of high-quality labeled data. In endodontics, segmenting the root pulp canal is notoriously difficult; these structures are curvilinear, extremely narrow, and exhibit low contrast against the surrounding dentin, often branching into complex topologies that are arduous to trace manually~\cite{RefinedPulp_UNet_2021}. Consequently, obtaining large-scale, voxel-level ground truth annotations for root canals is prohibitively expensive. Similarly, developing robust registration algorithms to fuse CBCT and IOS data requires diverse datasets of matched scans. Creating ground-truth alignment transformations is labor-intensive and requires expert verification, leading to a bottleneck in training fully supervised models for multi-modal registration~\cite{Kim2023_AutoReg_CBTC_IOS_segmentation}.

Semi-supervised learning (SSL) offers a potent solution to these limitations by leveraging limited labeled data alongside vast quantities of unlabeled images~\cite{SystematicReview_CBCT_IOS_registration_2025}. In the context of dental imaging, SSL can potentially learn robust anatomical representations of root canals and alignment features from unlabeled cohorts, reducing the reliance on extensive expert annotation. Yet, applying SSL to these specific dental tasks introduces unique challenges. For root canal segmentation, the model must handle extreme class imbalance between the tiny pulp volume and the background, while preventing the propagation of pseudo-label noise in disconnected structures~\cite{3DUNet_GlobalLocal_Loss_RootCanal2021}. For registration, the network must learn to align modalities with fundamentally different physical properties (optical surface vs. X-ray volume) without overfitting to a small set of aligned pairs~\cite{Elgarba2024_AI_CBTC_IOS_registration}.

\begin{table}[ht]
\centering
\caption{Summary of labeled and unlabeled data statistics for the Interoral Scanning (IOS) and Cone-Beam Computed Tomography (CBCT) datasets, including sample counts, size, image resolution, and other significant features.}
\label{tab:dataset_statistics}
\setlength{\tabcolsep}{5mm}
\renewcommand\arraystretch{1.3}
\resizebox{\linewidth}{!}{
\begin{tabular}{llcc}
\toprule
\textbf{Dataset} & \textbf{Metric} & \textbf{Labeled} & \textbf{Unlabeled} \\
\midrule
\multirow{5}{*}{IOS Part}
    & Sample Number           & 60     & 640 \\
    & Upper Jaw Number        & 30     & 320 \\
    & Lower Jaw Number        & 30     & 320 \\
    & Average Mesh Resolution (Vertices) & $\approx 1315435$ & $\approx 1209164$ \\
    & Average Mesh Resolution (Faces)    & $\approx 438815$  & $\approx 403271$ \\
\midrule
\multirow{7}{*}{CBCT Part}
    & Samples                 & 30     & 250 \\
    & Number of Slices        & 29240  & 60000 \\
    & Resolutions(Voxel)      & $512 \times 512$ / $445 \times 445$ & $281 \times 281$ \\
    & Matrix Size Range       & $266^2 - 640^2$   & $266^2 - 768^2$ \\
    & Voxel Size Range ($mm^3$) & $0.10 - 0.40$   & $0.08 - 0.45$ \\
    & HU Value Range          & $-1024 - 0$ / $-1000 - 0$ & $-1024 - 0$ / $-1000 - 0$ \\
    & Slice Thickness (mm)    & $0.3$ / $0.25$    & $0.3$ \\
\bottomrule
\end{tabular}
}
\end{table}

Despite the proliferation of dental image analysis research, existing public benchmarks have largely focused on tooth instance segmentation, neglecting internal pulp anatomy and multi-modal fusion. While previous initiatives successfully established baselines for semantic and instance-level tooth segmentation, they did not address the fine-grained segmentation of root canal systems or the geometric registration of IOS and CBCT data. The lack of standardized benchmarks for these complex tasks significantly impedes the translation of AI research into clinical applications.

To address these critical gaps, we organized the 3rd Semi-supervised Teeth Segmentation and Registration (STSR 2025) Challenge as an official satellite event of MICCAI 2025. STSR 2025 aims to benchmark state-of-the-art SSL algorithms on two clinically vital tasks. \textbf{Task 1} focuses on \textit{Semi-supervised Teeth and Pulp Root Canal Segmentation} in 3D CBCT. \textbf{Task 2} introduces \textit{Semi-supervised Teeth Crown and Root Registration}, challenging participants to align high-resolution IOS crown surfaces with volumetric CBCT root data.

In summary, the main contributions of this work are as follows:
\begin{itemize}
    \item We present a novel public dataset, the first large-scale, multi-modal resource specifically curated for semi-supervised root canal segmentation and CBCT-IOS registration.
    \item We established the first open international benchmark for these advanced dental tasks, employing a rigorous evaluation framework with Dockerized submissions.
    \item We provide a detailed analysis of SSL strategies employed by top-performing participants, offering valuable insights for future research.
\end{itemize}

\begin{table*}[ht]
\centering
\caption{Comparison of the \textbf{STS2025} dataset with CjData and other public 3D dental CBCT datasets. STS2025 introduces a significant semi-supervised benchmark with detailed fine-grained annotations for root canal structures.}
\label{tab:dental_cbct_datasets_updated}
\setlength{\tabcolsep}{2mm} 
\renewcommand\arraystretch{1.3}
\resizebox{\linewidth}{!}{ 
\begin{tabular}{lllllllcl} 
\toprule
\textbf{Dataset} & \textbf{Years} & \makecell[c]{\textbf{Age} \\ \textbf{Range}} & \textbf{Patients} & \makecell[c]{\textbf{Volumes} \\ \textbf{(A / woA)}} & \textbf{Slices} & \makecell[c]{\textbf{Num of} \\ \textbf{Teeth}} & \makecell[c]{\textbf{Annotator}} & \makecell[c]{\textbf{Volumes} \\ \textbf{(pixels)}} \\
\midrule

\makecell[l]{Clinical dental \\ CBCT~\cite{9083982}}
           & 2020 & 10-49      & 25  & \makecell[c]{25 \\ (25 / 0)}             & 9400   & $\approx$ 770   & 1  & \makecell[c]{$110\times145\times280$} \\

CTooth+\cite{cui2022ctooth+}                       & 2022 & 10--15 & 22  & \makecell[c]{168 \\ (22 / 146)} & 31380  & $\approx$ 5040  & 15  & \makecell[c]{$266\times266\times266$} \\

\makecell[l]{Mandibular Canal \\ CBCT~\cite{9686728}}
  & 2022 & 10-100      & 347   & \makecell[c]{347 \\ (347 / 0)}           & 88832       & $\approx$ 17220 & -  & \makecell[c]{$178\times423\times463$} \\

\makecell[l]{Multi-modal \\ dataset~\cite{huang2024multimodal}}
             & 2023 & 18-89      & 169   & \makecell[c]{188 \\ (188 / 188)} & 16203    & $\approx$ 5401  & 13 & \makecell[c]{$640\times640\times200$} \\

STS2023~\cite{wang2025miccai} & 2023 & 7--70  & 584 & \makecell[c]{584 \\ (84 / 500)}& 88500  & $\approx$ 17520  & 30 & \makecell[c]{$640\times640\times399$} \\
ToothFairy2\cite{bolelli2024segmenting} & 2023 & 16-100 & 530 & \makecell[c]{530\\(480/50)} & 80000 & $\approx$ 71400 & 5 & \textbf{\makecell[c]{$170\times272\times345$\\$298\times512\times512$}} \\

STS2024 & 2024 & 7--70  & 330 & \makecell[c]{330 \\ (30 / 300)} & 69960  & $\approx$ 9900  & 30 & \makecell[c]{$266\times266\times200$\\$512\times512\times332$} \\

ToothFairy3\cite{bolelli2025segmenting} & 2024 & 16-100 & 532 & \makecell[c]{532\\(532/0)} & 80000 & $\approx$ 17024 & 5 & \textbf{\makecell[c]{$170\times272\times345$\\$298\times512\times512$}} \\

\textbf{STS2025 (Ours)} & \textbf{2025} & \textbf{7--70}  & \textbf{370} & \makecell[c]{\textbf{370} \\ \textbf{(70 / 300)}} & - & - & \textbf{30+} & \makecell[c]{\textbf{Multi-resolution}} \\

\bottomrule
\end{tabular}
}
\end{table*}

\begin{table*}[t]
\centering
\caption{Comparison of the \textbf{STSR 2025} dataset with other public dental datasets. The STSR 2025 dataset provides a unique resource for \textbf{multi-modal registration}, featuring a large unlabeled set for semi-supervised learning. ``L / U'' denotes data with/without annotations (Labeled / Unlabeled).}
\label{tab:comparison}

\resizebox{\linewidth}{!}{%
\begin{tabular}{lcccccc}
\toprule

\textbf{Dataset} & \textbf{Year} & \textbf{Modality} & \textbf{Scans (L / U)} & \textbf{Num. of Teeth} & \textbf{Task} & \textbf{Precision / Res.} \\ 
\midrule

Zanjan\cite{zanjani2019deep} & 2019 & 3D Point Cloud & \begin{tabular}[c]{@{}c@{}}60 \\ (60 / 0)\end{tabular} & $\approx$ 3,700 & Semantic Seg. & $20~\mu m$ \\ \addlinespace

Teeth3DS \cite{ben2022teeth3ds} & 2022 & 3D IOS (Mesh) & \begin{tabular}[c]{@{}c@{}}1,800 \\ (1800 / 0)\end{tabular} & $\approx$ 23,999 & Instance Seg. & $10 \sim 90~\mu m$ \\ \addlinespace

3D-IOSSeg \cite{li2024fine} & 2024 & 3D IOS (Mesh) & \begin{tabular}[c]{@{}c@{}}440 \\ (440 / 0)\end{tabular} & $\approx$ 14,080 & Fine-grained Seg. & $10 \sim 80~\mu m$ \\ \addlinespace


\textbf{STSR 2025 (Ours)} & \textbf{2025} & \textbf{CBCT + IOS} & \textbf{\begin{tabular}[c]{@{}c@{}}330 \\ (30 / 300)\end{tabular}} & - & \textbf{Registration} & - \\ 
\bottomrule
\end{tabular}%
}
\end{table*}

\begin{figure*}[!ht]
\centering
\includegraphics[width=\linewidth]{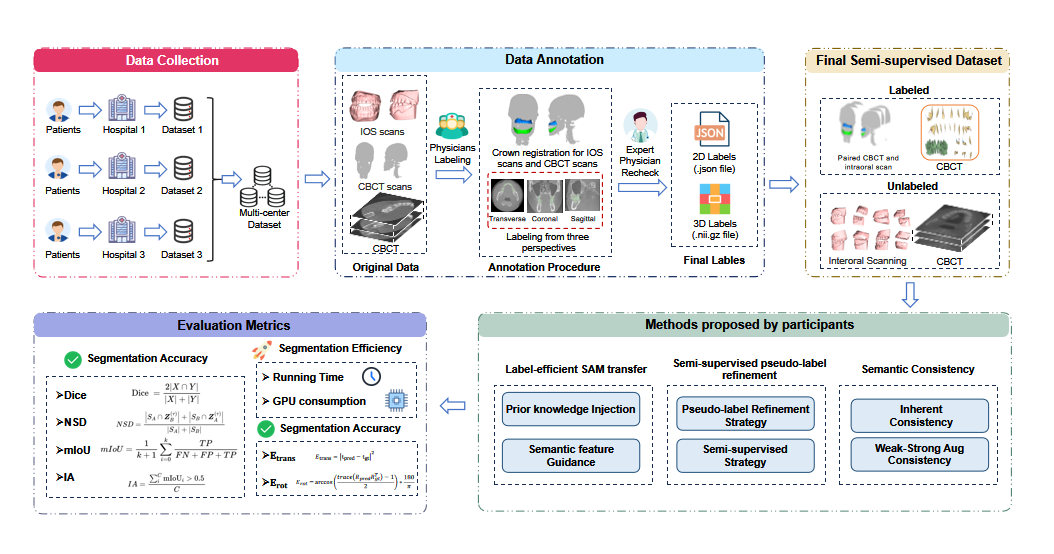}
\caption{End-to-end workflow of the MICCAI 2024 Semi-supervised Teeth Segmentation (STS) Challenge. The process encompasses five key stages: (1) multi-center data collection to ensure dataset diversity, (2) iterative annotation by clinicians for high-quality ground truth, (3) construction of the semi-supervised dataset with distinct labeled and unlabeled sets, and (4/5) the final summarization and evaluation of submitted participant methods.}
\label{fig:framework}
\end{figure*}

\section{Related Works}
\subsection{Semi-Supervised Dental Segmentation}
Accurate delineation of dental anatomical structures plays a fundamental role in orthodontic assessment, implant planning, endodontic navigation, and surgical simulation. Early studies on tooth segmentation from CBCT or panoramic radiographs primarily focused on distinguishing teeth from surrounding tissues, and benefited substantially from convolutional architectures such as U-Net, V-Net, and subsequent variants that improved 2D/3D medical image segmentation performance [1–3]. 3D U-Net–based models have shown particular effectiveness in CBCT by leveraging volumetric context to separate individual teeth from bone and soft tissue [4].

Beyond semantic segmentation, instance-level tooth segmentation has gained significant attention. Several works have introduced 3D instance segmentation pipelines capable of identifying and isolating each tooth despite metal artifacts, structural overlap, or irregular spacing. For example, Zanjani et al. developed a fully automated CBCT pipeline combining segmentation and identification [5], while Wang et al. further introduced a pose-aware instance segmentation model demonstrating robustness in challenging CBCT scenarios [6]. Nevertheless, most datasets emphasize crowns or full-tooth volumes; annotations for fine-scale internal structures—particularly the pulp chamber and multi-root canal systems—remain limited.

Segmentation of root canals and mandibular canals presents additional challenges due to their thin geometry, variable curvature, low contrast, and susceptibility to CBCT artifacts. Classical image-processing approaches such as level sets or vesselness filtering often fail to capture these structures reliably. Recent deep learning methods have demonstrated stronger performance, including hybrid level-set–constrained 3D U-Nets achieving Dice scores exceeding 0.95 [7], Attention U-Net–based systems for canal measurement [8], and two-stage models for mandibular canal extraction [9]. Nonetheless, these methods rely heavily on dense expert annotations, making large-scale supervised training impractical.

To reduce annotation burdens, semi-supervised learning (SSL) has been increasingly explored in medical imaging. Popular SSL approaches include pseudo-labeling [10], consistency-based teacher–student models [11], and perturbation-based regularization frameworks such as MixMatch or FixMatch [12,13]. While SSL has been applied to tooth segmentation on limited CBCT datasets, its use for fine-grained structures (pulp chambers, root canals) remains rare. The scarcity of publicly available CBCT datasets with detailed canal annotations further limits progress. A cohesive, hierarchical benchmark integrating tooth instance segmentation with canal-level SSL therefore remains unresolved.

\subsection{IOS–CBCT Registration and Multimodal Integration}
Accurate registration between intraoral scans (IOS) and CBCT has become essential in digital dentistry, enabling integration of high-resolution crown surfaces with volumetric root and bone anatomy. Traditional multimodal registration relies on geometric alignment techniques such as Iterative Closest Point (ICP) or curvature-based surface matching [14]. However, due to inherent modality differences—IOS lacking root geometry while CBCT provides internal structures—classical approaches often struggle with missing regions, noise, and field-of-view inconsistencies.

Learning-based frameworks have attempted to overcome these limitations by combining segmentation, identification, and registration within unified pipelines. Jang et al. proposed a fully automated CBCT–IOS integration workflow using tooth segmentation and global-to-local alignment strategies, followed by stitching-error correction for IOS mosaics [15]. These methods allow simultaneous incorporation of crowns, gingival surfaces, and roots into a unified representation.

More recent clinical studies have demonstrated that AI-driven multimodal registration can surpass manual or semi-automated methods in both accuracy and efficiency. One validated system achieved median surface deviations as low as 0.04 mm and root-mean-square registration errors of 0.19 mm across 31 patients [16], while another multicenter evaluation reported that automated registration was over seven times faster than expert-guided registration with comparable precision [17]. These findings highlight the feasibility of robust, clinically deployable dental registration systems.

Despite this progress, critical gaps remain: current datasets rarely include standardized crown-root correspondences across CBCT and IOS, and few incorporate detailed segmentation (crowns, roots, canals) alongside registration annotations. Such limitations hinder fair benchmarking across algorithms and restrict generalization to diverse clinical conditions. Developing integrated datasets and standardized evaluation protocols—covering semi-supervised canal segmentation and multimodal IOS–CBCT fusion—remains a key step toward comprehensive and reliable 3D dental modeling frameworks.

\begin{figure*}[!ht]
\centering
\includegraphics[width=\linewidth]{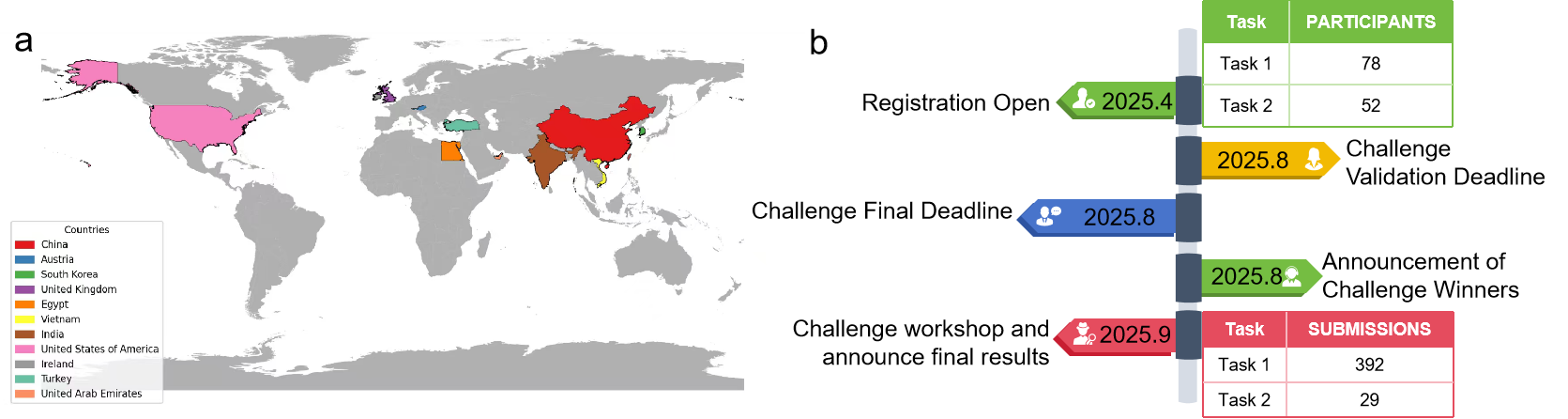}
\caption{Overview of the STS 2025 Challenge participation and timeline. (a) World map illustrating the geographical distribution of registered teams. (b) Detailed timeline of the challenge schedule from registration and training to the final workshop and announcement of results, together with the number of participating teams (78 for Task 1 and 52 for Task 2) and the total submissions (392 for Task 1 and 29 for Task 2).}
\label{fig:schedule}
\end{figure*}

\section{Challenge Description}
\subsection{Dataset information and annotation details}
The STSR (Semi-supervised Teeth Segmentation and Registration) 2025 Challenge is one of the official competitions at the MICCAI 2025 ONDI Workshop. It primarily aims to systematically evaluate, in real clinical scenarios, semi-supervised learning-based methods for 3D tooth and root canal segmentation, as well as CBCT–IOS crown–root registration techniques. Unlike STS 2024, which focused solely on 2D panoramic and 3D CBCT tooth instance segmentation, this challenge further emphasizes the precise segmentation of 3D CBCT tooth crowns and pulp root canals, as well as multi-modal crown-root registration between CBCT and IOS. It specifically evaluates algorithm performance in multi-instance, multi-class, and multi-modal scenarios. The challenge comprises two tasks: Task 1 involves semi-supervised segmentation of 3D CBCT tooth crowns and root canals, while Task 2 addresses semi-supervised crown-root registration between 3D CBCT and IOS data.

The challenge dataset originates from real clinical settings, jointly provided by multiple collaborating dental institutions. It encompasses diverse clinical scenarios including routine endodontic treatment, orthodontic therapy, and implant restoration. All CBCT images were acquired by radiologists or dentists with over five years of experience following standardized examination protocols. IOS data originated from intraoral 3D scans of the same patients within a similar time window. Cases extensively featured dental malocclusion, tooth loss, root canal treatment traces, metal restorations, orthodontic brackets, implants, and various X-ray artifacts, fully reflecting the complexity of real clinical images. All data undergoes standardized quality control and anonymization prior to inclusion, removing personally identifiable metadata while retaining only imaging parameters and essential identifiers relevant for research and evaluation. Data usage is approved by the corresponding medical ethics committee and governed by a CC BY-NC-ND license, permitting solely scientific research purposes. Commercial development and secondary distribution in any form are strictly prohibited.

\subsubsection{3D-CBCT Dataset}
Task 1: For 3D CBCT volumetric data, perform precise three-dimensional segmentation of all permanent teeth (including wisdom teeth) and their root canal pulp chambers. Unlike previous approaches that only annotated tooth contours or crowns, this task further annotates internal root canal structures on top of tooth segmentation. This enables the model to reconstruct crowns, roots, and root canal spaces simultaneously, providing more detailed anatomical information for applications such as root canal therapy, complex extractions, and implant planning. All CBCTs undergo standardized reconstruction and resampling to normalize voxel resolution and coordinate systems. An two-stage annotation process is employed: initial masks are drawn by trained intermediate dentists using ITK-SNAP, followed by case-by-case refinement and arbitration by experts with extensive clinical experience. Each tooth and its root canal is assigned a unique ID, enabling instance-level 3D annotation integrated with FDI tooth position coding. The training set comprises 30 precisely annotated CBCTs and approximately 300 unannotated cases. The validation set contains 40 annotated CBCTs for online evaluation. The test set consists of annotated CBCTs visible only to the organizing committee for final ranking.

\subsubsection{CBCT–IOS Dataset}
Task 2 focuses on crown-root registration between 3D CBCT and intraoral scanners (IOS), leveraging the complementary strengths of IOS's high-resolution crown surface capture and CBCT's comprehensive crown-root-jawbone information. IOS models offer exceptional resolution for crown surface geometry but lack root and jawbone data; CBCT provides comprehensive three-dimensional anatomical information encompassing the crown, root, and surrounding bony structures, but faces limitations in spatial resolution and artifact generation. Each sample consists of a pair of registered CBCT volumetric data and IOS dental arch models. IOS undergoes preprocessing including coordinate normalization and mesh cleaning. Registration ground truth is obtained by manually fine-tuning a coarse algorithm-based registration in a 3D environment by clinicians and engineers, yielding one set of 4×4 rigid transformation matrices for each jaw. Data division: The training set contains 30 cases with precise registration annotations and approximately 300 unannotated paired data sets. The validation set includes 50 paired data sets with ground truth matrices for online evaluation. The test set comprises hidden paired samples visible only to the organizing committee, used for final ranking.

\subsection{Participants and challenge phases}
The STS2025 (STSR 2025) Challenge aims to build upon previous work and further advance the research and practical application of semi-supervised learning methods in medical and dental image analysis. The competition will proceed in phases: Registration opens on April 25, 2025, with the release of training data and the commencement of eligibility verification based on the Data Access Agreement. Teams that pass verification will receive data download access and detailed usage instructions. To facilitate early algorithm debugging, a non-scored test set will be released on April 28. Participants must submit results for the scored test set by August 15, with Docker submissions due by August 22. Each team is typically allowed only one valid submission during Test Set Phase B to mitigate overfitting risks. The organizing committee will conduct evaluations in a standardized environment and announce the final award list on August 31. Teams may prepare and submit technical reports between September 1–14, with acceptance notifications issued following review. Challenge outcomes will be showcased at a dedicated workshop during MICCAI 2025 (September 23–27, 2025). Final papers must be proofread and submitted to the conference proceedings by October 8.

Tasks 1 and 2 attracted research teams from multiple countries and regions worldwide, with China contributing the largest number of participating teams. Participation statistics indicate that Task 1 (Segmentation of Teeth and Pulp Canals in CBCT Scans) attracted 78 teams submitting a total of 392 solutions, while Task 2 (Crown-Root Registration in IOS and CBCT Scans) drew 52 teams submitting 29 solutions. These figures demonstrate sustained global interest in dental image segmentation technology and its modeling and application within semi-supervised learning frameworks. To recognize outstanding contributions, both tracks adopted a unified reward system: the top two teams on the test leaderboard received cash prizes ($300 for first place, $100 for second place) along with honorary certificates.

To foster collaboration and reproducibility, the organizing committee provided official baseline code based on nnU-Net along with corresponding Docker images during the competition. In partnership with cloud platforms, GPU computing subsidies were offered to registered teams, lowering the hardware barrier for training large-scale 3D models. During the online validation phase, participants may submit segmentation or registration result files on the Codabench platform for immediate feedback. Alternatively, they may package their complete algorithms via Docker for evaluation by the organizing committee in a standardized environment. Following the conclusion of the STS2025 Challenge, the first authors of the top five teams on the leaderboard will be invited to co-author a challenge summary review paper. This paper will systematically outline the task specifications, methodological features, and experimental results, with plans to submit it to a top-tier medical imaging journal. Additionally, all teams are encouraged to submit technical reports. After peer review, these reports will be included in conference proceedings such as Springer LNCS, further advancing the exchange and practical application of related methods.

\subsection{Clinical Utility of Segmentation}
Both tasks in this year's STSR 2025 Challenge possess clear and significant clinical applications. For Task 1, precise 3D segmentation of tooth structures and root canals serves as a critical foundation for various clinical procedures, including root canal therapy, complex extractions, implant restorations, and orthodontic treatments. Fine-grained segmentation of tooth structures and root canal spaces within CBCT images enables clear visualization of root canal number, orientation, branching, and curvature. This assists clinicians in accurately assessing preoperative root canal preparation difficulty and filling strategies, reducing risks of complications such as perforation and residual canals. It also holds significant value for postoperative efficacy evaluation and quantitative analysis of lesion progression during long-term follow-up.
For Task 2, CBCT–IOS crown–root registration integrates high-resolution crown surface geometry with CBCT volumetric data encompassing roots, alveolar bone, and critical surrounding anatomy (e.g., mandibular canal, maxillary sinus) within a unified 3D coordinate system. In implant planning, multimodal data based on precise registration assists clinicians in designing the three-dimensional position and inclination of implants, enabling comprehensive assessment of bone volume, bone quality, and safe distances from critical anatomical structures in the implant site. In orthodontic treatment, registration results facilitate combined analysis of crown morphology, root position, and alveolar bone coverage, thereby establishing safer and more effective tooth movement pathways. In maxillofacial and orthognathic surgery, virtual surgical planning and postoperative realignment assessment based on registered 3D models hold significant application value. Overall, instance-level 3D segmentation and multimodal registration advance traditional digital oral technologies into the era of intelligent diagnosis and precision treatment.

\subsection{Performance Evaluation}
For this challenge, we designed a systematic quantitative evaluation metric system for two tasks to comprehensively characterize segmentation accuracy, registration accuracy, and inference efficiency. For Task~1 (3D CBCT tooth and root canal segmentation), we measure mask overlap and boundary alignment at both the image level and instance level. Let the predicted mask of the entire image be $A$ and the ground truth mask be $B$. The image-level Dice coefficient is defined as
\begin{equation*}
    \mathrm{Dice}_{\text{image}}
    = \frac{2\lvert A \cap B \rvert}{\lvert A \rvert + \lvert B \rvert} \, ,
\end{equation*}
and the image-level mean Intersection over Union (mIoU) is defined as
\begin{equation*}
    \mathrm{mIoU}_{\text{image}}
    = \frac{\lvert A \cap B \rvert}{\lvert A \cup B \rvert} \, .
\end{equation*}

At the instance level, let the predicted mask and ground truth mask for the $i$-th instance (tooth or root canal) be $A_i$ and $B_i$, respectively, with a total of $N$ instances. The Dice and mIoU for this instance are given by
\begin{equation*}
    \mathrm{Dice}_i
    = \frac{2\lvert A_i \cap B_i \rvert}{\lvert A_i \rvert + \lvert B_i \rvert} \, ,
    \qquad
    \mathrm{mIoU}_i
    = \frac{\lvert A_i \cap B_i \rvert}{\lvert A_i \cup B_i \rvert} \, .
\end{equation*}
Averaging across all instances yields the instance-level Dice and mIoU metrics:
\begin{equation*}
    \mathrm{Dice}_{\text{instance}}
    = \frac{1}{N} \sum_{i=1}^{N} \mathrm{Dice}_i ,
    \qquad
    \mathrm{mIoU}_{\text{instance}}
    = \frac{1}{N} \sum_{i=1}^{N} \mathrm{mIoU}_i \, .
\end{equation*}

Relying solely on voxel overlap is insufficient to comprehensively reflect boundary errors. Therefore, we further employ Normalized Surface Dice (NSD) to measure the surface-level agreement between predictions and ground truth. Given a distance tolerance $\tau$ (e.g., $2\,\mathrm{mm}$), let $\mathrm{overlap}_{\text{GT}}$ and $\mathrm{overlap}_{\text{pred}}$ denote the surface areas of overlap between ground truth and predicted boundaries within tolerance $\tau$ at the image level, and let $\mathrm{total\_area}_{\text{GT}}$ and $\mathrm{total\_area}_{\text{pred}}$ denote the total surface areas of the ground truth and predicted boundaries, respectively. The image-level NSD is defined as
\begin{equation*}
    \mathrm{NSD}_{\text{image}}
    =
    \frac{
        \mathrm{overlap}_{\text{GT}} + \mathrm{overlap}_{\text{pred}}
    }{
        \mathrm{total\_area}_{\text{GT}} + \mathrm{total\_area}_{\text{pred}}
    } \, .
\end{equation*}
For the $i$-th instance, similarly denote $\mathrm{overlap}_{\text{GT}_i}$, $\mathrm{overlap}_{\text{pred}_i}$, $\mathrm{total\_area}_{\text{GT}_i}$, and $\mathrm{total\_area}_{\text{pred}_i}$. The instance-level NSD is then defined as
\begin{equation*}
    \mathrm{NSD}_i
    =
    \frac{
        \mathrm{overlap}_{\text{GT}_i} + \mathrm{overlap}_{\text{pred}_i}
    }{
        \mathrm{total\_area}_{\text{GT}_i} + \mathrm{total\_area}_{\text{pred}_i}
    } \, ,
\end{equation*}
and averaging across all instances yields
\begin{equation*}
    \mathrm{NSD}_{\text{instance}}
    = \frac{1}{N} \sum_{i=1}^{N} \mathrm{NSD}_i \, .
\end{equation*}

At the instance recognition level, we employ the Instance Agreement (IA) metric to evaluate the model's overall recognition capability across different tooth positions. Let $\mathcal{C}_{\text{GT}}$ and $\mathcal{C}_{\text{pred}}$ denote the sets of categories appearing in the ground truth and predictions, respectively. If the IoU between a category $c$ in the prediction and the ground truth satisfies $\mathrm{IoU}(c) \ge 0.5$, then that category is considered successfully recognized. Define
\begin{equation*}
    \mathcal{C}_{\text{match}}
    =
    \bigl\{ c \,\big|\, \mathrm{IoU}(c) \ge 0.5 \bigr\} ,
    \qquad
    \mathcal{C}_{\text{all}}
    =
    \mathcal{C}_{\text{GT}} \cup \mathcal{C}_{\text{pred}} \, ,
\end{equation*}
then IA is defined as
\begin{equation*}
    \mathrm{IA}
    = \frac{
        \# \mathcal{C}_{\text{match}}
    }{
        \# \mathcal{C}_{\text{all}}
    } \, ,
\end{equation*}
where $\#(\cdot)$ denotes the cardinality of a set. The aforementioned image-level and instance-level Dice, mIoU, NSD, and IA collectively form the segmentation accuracy evaluation metrics for Task~1.

For Task~2 (CBCT--IOS crown--root registration), we primarily evaluate rigid geometric registration error. Let the ground truth rigid transformation and predicted rigid transformation be
\begin{equation*}
    T_{\text{gt}} =
    \begin{bmatrix}
        R_{\text{gt}} & t_{\text{gt}} \\
        0 & 1
    \end{bmatrix},
    \qquad
    T_{\text{pred}} =
    \begin{bmatrix}
        R_{\text{pred}} & t_{\text{pred}} \\
        0 & 1
    \end{bmatrix},
\end{equation*}
where $R_{\text{gt}}, R_{\text{pred}} \in \mathbb{R}^{3 \times 3}$ are rotation matrices, and $t_{\text{gt}}, t_{\text{pred}} \in \mathbb{R}^{3}$ are translation vectors. Translation error (TransErr) is defined as the Euclidean distance between the two translation vectors:
\begin{equation*}
    \mathrm{TransErr}
    = \bigl\| t_{\text{pred}} - t_{\text{gt}} \bigr\|_2 \, ,
\end{equation*}
which reflects the overall positional offset after registration. Rotation error (RotErr) is measured by the angle between the relative rotation matrices. First, we construct the relative rotation matrix
\begin{equation*}
    R_{\text{rel}}
    = R_{\text{pred}} \, R_{\text{gt}}^{\mathsf{T}} \, ,
\end{equation*}
and compute the relative rotation angle $\theta$ (in radians) from its trace:
\begin{equation*}
    \theta
    = \cos^{-1}
    \left(
        \frac{\mathrm{trace}(R_{\text{rel}}) - 1}{2}
    \right) .
\end{equation*}
Converting to degrees yields
\begin{equation*}
    \mathrm{RotErr}
    = \theta \cdot \frac{180}{\pi} \, .
\end{equation*}
Averaging across all cases and separately for maxilla and mandible yields the overall translation and rotation errors for Task~2.

To further analyze multimodal alignment quality from grayscale consistency and information-theoretic perspectives, we define the aligned CBCT volumetric data as $X$ and the IOS volumetric data as $Y$ in a unified coordinate system, where the voxel index is $i$ and the means are $\mu_X$ and $\mu_Y$, respectively. The normalized cross-correlation (NCC) is defined as
\begin{equation*}
    \mathrm{NCC}(X, Y)
    = \frac{
        \sum_{i} (X_i - \mu_X)(Y_i - \mu_Y)
    }{
        \sqrt{\sum_{i} (X_i - \mu_X)^2}
        \;\sqrt{\sum_{i} (Y_i - \mu_Y)^2}
    } \, .
\end{equation*}
Let entropy be denoted by $H(\cdot)$ and joint entropy by $H(X,Y)$. Then mutual information (MI) and normalized mutual information (NMI) are respectively given by
\begin{equation*}
    \mathrm{MI}(X, Y)
    = H(X) + H(Y) - H(X, Y) ,
\end{equation*}
\begin{equation*}
    \mathrm{NMI}(X, Y)
    = \frac{H(X) + H(Y)}{H(X, Y)} \, .
\end{equation*}

In terms of efficiency evaluation, both tasks employ a unified metric system. Let there be $M$ cases in total, with the inference time for the $k$-th case denoted as $\mathrm{RT}_k$. The average inference time is defined as
\begin{equation*}
    \overline{\mathrm{RT}}
    = \frac{1}{M} \sum_{k=1}^{M} \mathrm{RT}_k ,
\end{equation*}
and each case's inference time must not exceed the preset upper limit $T_{\max}$ (e.g., $60\,\mathrm{s}$), i.e.,
\begin{equation*}
    \mathrm{RT}_k \le T_{\max}, \qquad k = 1,\dots,M \, .
\end{equation*}
Regarding GPU memory consumption, we sample the GPU memory usage $m(t_i)$ at fixed intervals during inference. The area under the memory--time curve (AUC-GPU) can be approximated as
\begin{equation*}
    \mathrm{AUC\!-\!GPU}
    \approx
    \sum_{i=1}^{n-1}
        m(t_i)\,\bigl(t_{i+1} - t_i\bigr) \, .
\end{equation*}
This metric simultaneously accounts for both peak memory usage and the duration of high memory occupancy, providing a comprehensive measure of memory overhead. By jointly employing the aforementioned segmentation accuracy, registration accuracy, and efficiency metrics, we can systematically evaluate the accuracy, stability, and practical usability of competing methods across multiple dimensions.

\section{Methods}
\begin{figure*}[!ht]
\centering
\includegraphics[width=\linewidth]{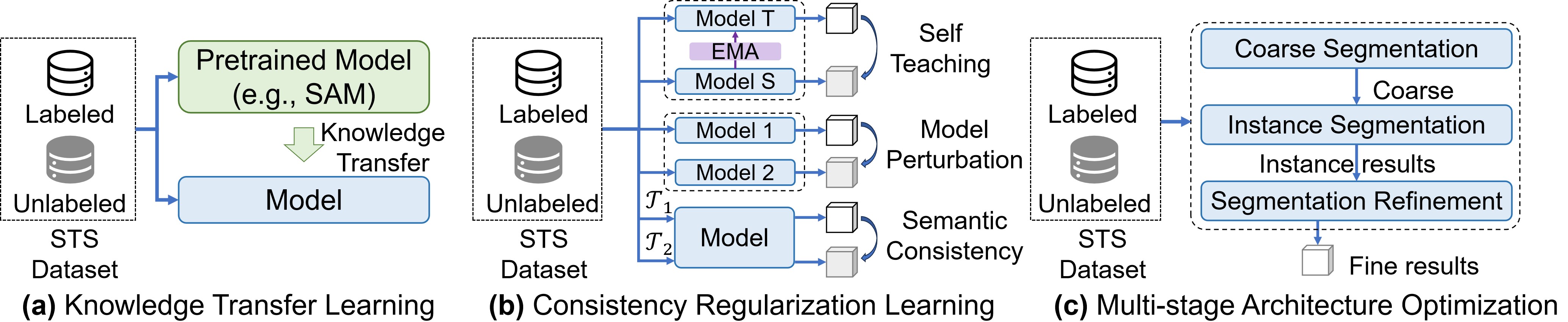}
\caption{Overview of prominent methodological strategies employed by participants in the STS 2024 Challenge. The figure illustrates four key approaches: (a) Knowledge transfer with pretrained models, where pre-trained foundation models (e.g., SAM) are leveraged to improve segmentation. (b) Consistency regularization learning, including self-teaching, model perturbation, and semantic consistency ($\mathcal{T}_{1}$ and $\mathcal{T}_{2}$ denote two kinds of transformation). (c) Multi-stage architecture optimization decomposes the problem into multiple sub-problems and gradually obtained fine results.}
\label{fig:method1}
\end{figure*}

\begin{table*}[t!]
    \centering
    \caption{Summary of 2D submitted teams, model architecture, optimization strategy, and training settings.}
    \label{tab:team_summary}
    \renewcommand{\arraystretch}{1.3} 
    \setlength{\tabcolsep}{4pt}
    \begin{adjustbox}{width=\textwidth}
    \begin{tabular}{lccccccc}
        \toprule
        \textbf{Team} & \makecell[c]{\textbf{Model}\\\textbf{Architecture}} & \textbf{Backbone} & \textbf{Optimizer} & \makecell[c]{\textbf{Loss}\\\textbf{Function}} & \textbf{Device} & \textbf{Epochs} & \makecell[c]{\textbf{Batch}\\\textbf{Size}} \\
        \midrule
        ChohoTech & YOLOv8 & \makecell[c]{YOLOv8} & Adam & \makecell[c]{CIoU, DFL,\\VFL} & RTX 3090 & 100 & 32 \\ \midrule
        Camerart2024 & \makecell[c]{Self-Training\\Pipeline} & DeepLabV3+ & Adam & BCE, Dice & RTX 4090 & 200 & 4 \\ \midrule
        Jichangkai & \makecell[c]{Two-Stage\\Semi-Supervised\\nnU-Net} & nnU-Net & AdamW & Dice, CE & RTX 4090 & 150 & 4 \\ \midrule
        Dew123 & DICL Network & UNet & SGD & \makecell[c]{Dice, MSE,\\CE} & RTX 3060 & 100 & 8 \\ \midrule
        Junqiangmler & Semi-TeethSeg2024 & VNet2d & AdamW & \makecell[c]{Dice,\\Cross-Entropy} & RTX 4090 & 300 & 4 \\  \midrule 
        Isjinghao & SemiT-SAM & SAM & AdamW & \makecell[c]{Multi-\\component} & RTX 3060 & 300 & 4 \\ \midrule 
        Lazyman & \makecell[c]{Cross Teaching\\Network} & \makecell[c]{CNN +\\Transformer} & SGD & Dice, MSE & RTX 4090 & 43 & 16 \\ \midrule 
        Caiyichen & YOLOv9 & \makecell[c]{YOLOv9} & Adam & \makecell[c]{CIoU, DFL,\\VFL} & RTX 3060 & 100 & 16 \\ \midrule 
        Guo7777 & \makecell[c]{ResUnet50\\+ SAM} & \makecell[c]{ResNet50,\\SAM} & Adam & \makecell[c]{BCEWithLogitsLoss,\\MSELoss} & \makecell[c]{Tesla V100\\-SXM2} & 300 & 4 \\ \midrule 
        Ccc2024 & DAE-Net & \makecell[c]{Dual Attention\\Mechanism} & Adam & Dice, IoU & RTX 4060 Ti & 40 & 32 \\
        \bottomrule
    \end{tabular}
    \end{adjustbox}
\end{table*}

\begin{table*}[t!]
    \centering
    \caption{Summary of 3D submitted teams, model architecture, optimization strategy, and training settings for STS MICCAI 2024 Challenge Task 2.}
    \label{tab:team_summary_3d_combined_final}
    \renewcommand{\arraystretch}{1.3} 
    \setlength{\tabcolsep}{4pt} 
    \begin{adjustbox}{width=\textwidth}
    \begin{tabular}{llccccc} 
        \toprule
        \textbf{Team} & \makecell[c]{\textbf{Model}\\\textbf{Architecture}} & \textbf{Backbone} & \textbf{Optimizer} & \makecell[c]{\textbf{Loss}\\\textbf{Function}} & \textbf{Device} & \makecell[c]{\textbf{Epochs}}\ \\
        \midrule
        Chohotech & \makecell[c]{3-Stage\\Pipeline} & \makecell[c]{YOLOv8,\\U-Net} & Adam & \makecell[c]{CIoU, DFL,\\VFL} & NVIDIA A100 & 100 \\
        \midrule
        Houwentai & \makecell[c]{CFP 2-Stage\\Semi-sup.\\nnU-Net} & \makecell[c]{nnU-Net} & AdamW & Dice, CE & 6 $\times$ RTX 4090 & 100 \\
        \midrule
        Madongdong & \makecell[c]{Semi-supervised\\YOLOv8} & YOLOv8 & Adam & \makecell[c]{CIoU, DFL,\\VFL} & RTX 3090 & 300 \\
        \midrule
        Jichangkai & \makecell[c]{2-Stage nnU-Net\\Self-training} & nnU-Net & AdamW & Dice, CE & RTX 3090 & 300 \\
        \midrule
        Junqiangmler & \makecell[c]{ROI Preproc.\\+ VNet3d} & VNet3d & AdamW  & Dice, CE & RTX 4090 & 300 \\
        \midrule
        Gute\_iici & \makecell[c]{2-Stage\\Unimatch} & \makecell[c]{VNet (S1),\\Enc-Dec (S2)} & AdamW & Unimatch & Tesla V100 & 100  \\
        \bottomrule
    \end{tabular}
    \end{adjustbox}
\end{table*}

\subsection{Approaches based on the nnU-Net Framework}
The nnU-Net framework, renowned for its self-configuring pipeline and strong performance in medical imaging, served as a powerful baseline for several teams. These approaches focused their innovations not on redesigning the core architecture, but on implementing effective semi-supervised learning strategies and optimizing the framework for clinical efficiency.

\subsubsection{Ji Changkai}
This team proposed a solution centered on the 3D full-resolution nnU-Net architecture~\cite{isensee2021nnu}, with a primary focus on balancing high accuracy with computational efficiency. Their approach, termed Efficient nnU-Net, integrated a customized preprocessing pipeline and lightweight post-processing. The core innovation lay in two key inference acceleration strategies: the removal of computationally expensive test-time augmentation (TTA) and an optimized interpolation process using PyTorch's native functions. For training, they utilized a standard composite loss function of Dice and Cross-Entropy. This submission prioritized creating a model that was not only accurate but also practical for deployment in time-sensitive clinical environments.

\subsubsection{DiceMed}
This team based their solution on the nnU-Net framework, enhancing its performance through a direct and effective semi-supervised strategy. Their core contribution was a two-stage pseudo-labeling scheme. In the first stage, a baseline nnU-Net model was trained for 750 epochs exclusively on the 30 labeled scans. In the second stage, this model was used to generate pseudo-labels for the 300 unlabeled scans. The model was then retrained for an additional 500 epochs on the combined dataset of original labels and high-confidence pseudo-labels. This approach aimed to directly leverage the unlabeled data to expand the training set and improve the model's ability to recognize diverse anatomical variations.

\begin{figure}[!ht]
    \centering
    \includegraphics[width=\linewidth]{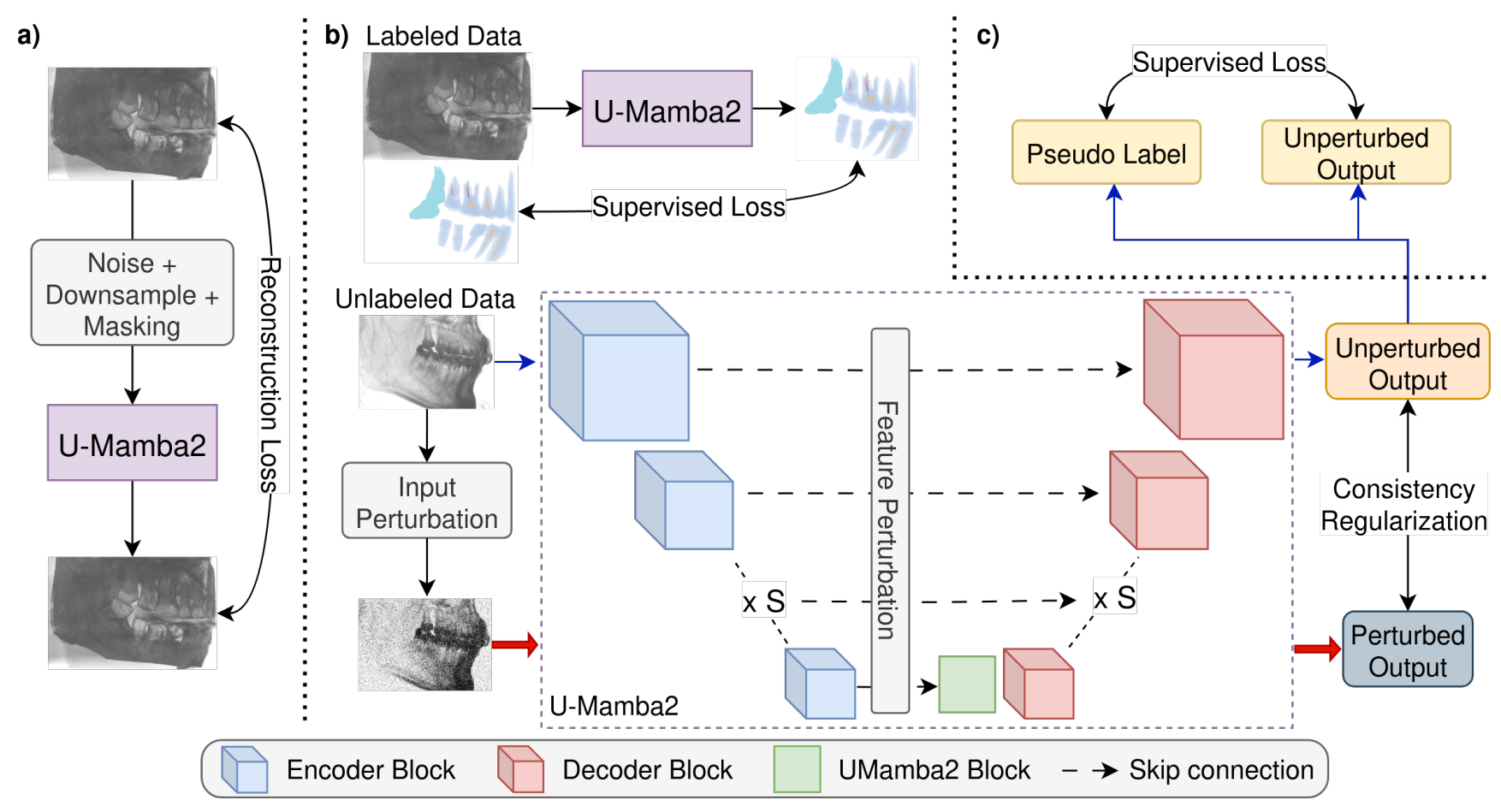}
    \caption{Overall diagram of the proposed U-Mamba2-SSL framework. (a) UMamba2 is first pre-trained by reconstructing inputs corrupted with noise, downsampled, and masked; (b) The second stage involves a combination of supervisedloss for the labeled data and consistency regularization between the unperturbedoutput and the perturbed output of the unlabeled data; (c) The final stage introduces pseudo labeling on top of the training objectives in (b). Only pseudo labels with confidence above a certain threshold contribute to the training loss.}
    \label{fig:2d_rank6}
\end{figure}

\subsection{Hybrid Architectures with State Space Models}
To address the inherent limitations of standard convolutional networks in modeling long-range spatial dependencies within large 3D volumes, several teams developed novel hybrid architectures. These methods integrate State Space Models (SSMs), such as Mamba, into a U-Net backbone to efficiently capture global context while retaining the U-Net's strength in hierarchical feature extraction.

\subsubsection{Zhi Qin}
This team introduced a novel architecture, U-Mamba2, which integrates the Mamba2 State Space Model into the bottleneck of a U-Net to enhance the modeling of global spatial relationships. Their methodology was distinguished by a sophisticated three-stage semi-supervised learning pipeline. The process began with self-supervised pre-training using a Denoising Autoencoder (DAE) on all data. This was followed by a consistency regularization stage, where the model was trained to produce stable outputs under both input and feature space perturbations. The final stage introduced a pseudo-labeling scheme that used a high confidence threshold (0.75) to generate reliable labels from the unlabeled set.

\begin{figure}[!ht]
    \centering
    \includegraphics[width=\linewidth]{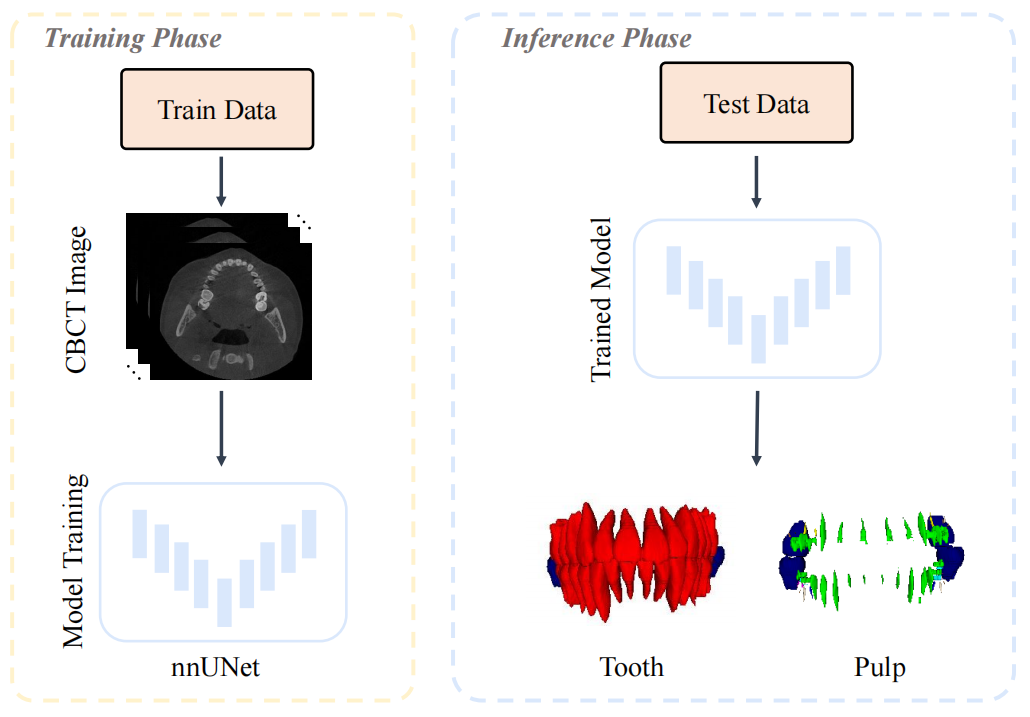}
    \caption{The overall structure of our nnU-Net-based model for teeth and root pulp canal
    segmentation.}
    \label{fig:2d_rank3}
\end{figure}

\subsubsection{Chen Liangyu}
This team proposed an innovative hybrid architecture named TCM-UNet. This model integrates several advanced components into a 3D U-Net backbone to improve feature representation. The encoder's core feature extractor is the CMambaBlock, a module with three distinct branches for explicit long-range context modeling (via SSM), complex feature extraction, and general feature learning. Furthermore, they introduced a Tri-Axis Attention Module (TAAM) into the skip connections to address the challenge of cross-dimensional feature correlation in 3D volumes. For training, they employed a composite Dice and Cross-Entropy loss function enhanced with a deep supervision mechanism to ensure robust learning across all levels of the decoder.

\begin{figure}[!ht]
    \centering
    \includegraphics[width=0.8\linewidth]{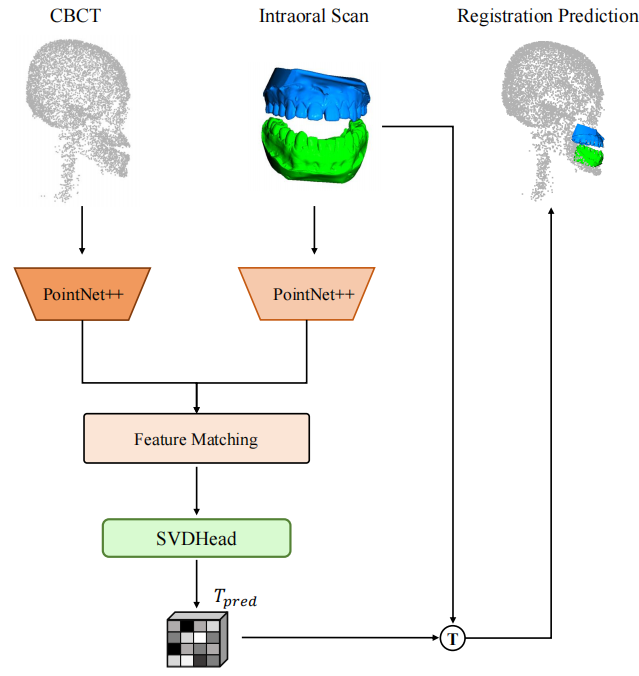}
    \caption{CBCT scans and intraoral scans are separately processed by PointNet++, followed by feature matching and rigid transformation estimation using SVD. The predicted transformation is applied to align the intraoral scans with CBCT data.}
    \label{fig:3d_rank6} 
\end{figure}

\section{Results}














\begin{table*}[t!]
    \centering
    \caption{Quantitative comparison of segmentation performance for ranked teams. The evaluation includes Image-level and Instance-level metrics (DSC, IoU, NSD, and IA). Values are presented as mean (standard deviation).}
    \label{table:results_2d}
    \renewcommand{\arraystretch}{1.4} 
    \begin{adjustbox}{width=\textwidth}
       \begin{tabular}{lc|ccc|cccc}
        \toprule
        \multicolumn{2}{c|}{\textbf{Team}} & \multicolumn{3}{c|}{\textbf{Image-level}} & \multicolumn{4}{c}{\textbf{Instance-level}}\\
        \midrule
        Name & Ranking 
        & \makecell[c]{DSC $\uparrow$} 
        & \makecell[c]{IoU $\uparrow$} 
        & \makecell[c]{NSD $\uparrow$} 
        & \makecell[c]{DSC $\uparrow$} 
        & \makecell[c]{IoU $\uparrow$} 
        & \makecell[c]{NSD $\uparrow$} 
        & \makecell[c]{IA $\uparrow$} \\
        \midrule

        zhiqin1998 & 1 & \makecell[c]{0.9176 \\ (\scriptsize{$\pm$}0.1970)} & \makecell[c]{0.8822 \\ (\scriptsize{$\pm$}0.1924)} & \makecell[c]{0.9487 \\ (\scriptsize{$\pm$}0.2042)} & \makecell[c]{0.6668 \\ (\scriptsize{$\pm$}0.1788)} & \makecell[c]{0.5513 \\ (\scriptsize{$\pm$}0.1589)} & \makecell[c]{0.8838 \\ (\scriptsize{$\pm$}0.2058)} & \makecell[c]{0.5778 \\ (\scriptsize{$\pm$}0.2589)} \\ 

        \midrule

        jichangkai & 2 & \makecell[c]{0.9431 \\ (\scriptsize{$\pm$}0.0183)} & \makecell[c]{0.8929 \\ (\scriptsize{$\pm$}0.0314)} & \makecell[c]{0.9934 \\ (\scriptsize{$\pm$}0.0089)} & \makecell[c]{0.6268 \\ (\scriptsize{$\pm$}0.0940)} & \makecell[c]{0.4994 \\ (\scriptsize{$\pm$}0.0824)} & \makecell[c]{0.8786 \\ (\scriptsize{$\pm$}0.1114)} & \makecell[c]{0.4748 \\ (\scriptsize{$\pm$}0.1690)} \\ 
        \midrule

        DiceMed & 3 & \makecell[c]{0.8478 \\ (\scriptsize{$\pm$}0.2397)} & \makecell[c]{0.7862 \\ (\scriptsize{$\pm$}0.2498)} & \makecell[c]{0.8530 \\ (\scriptsize{$\pm$}0.2623)} & \makecell[c]{0.6019 \\ (\scriptsize{$\pm$}0.1798)} & \makecell[c]{0.4815 \\ (\scriptsize{$\pm$}0.1532)} & \makecell[c]{0.8110 \\ (\scriptsize{$\pm$}0.2326)} & \makecell[c]{0.5099 \\ (\scriptsize{$\pm$}0.2486)} \\ 
        \midrule

        cccc2024 & 4 & \makecell[c]{0.4521 \\ (\scriptsize{$\pm$}0.4823)} & \makecell[c]{0.4344 \\ (\scriptsize{$\pm$}0.4636)} & \makecell[c]{0.4704 \\ (\scriptsize{$\pm$}0.5017)} & \makecell[c]{0.3052 \\ (\scriptsize{$\pm$}0.3315)} & \makecell[c]{0.2477 \\ (\scriptsize{$\pm$}0.2699)} & \makecell[c]{0.4156 \\ (\scriptsize{$\pm$}0.4502)} & \makecell[c]{0.2509 \\ (\scriptsize{$\pm$}0.2918)} \\ 
        \midrule

        minh\_dang & 5 & \makecell[c]{0.8105 \\ (\scriptsize{$\pm$}0.0904)} & \makecell[c]{0.6902 \\ (\scriptsize{$\pm$}0.1184)} & \makecell[c]{0.8666 \\ (\scriptsize{$\pm$}0.0915)} & \makecell[c]{0.1464 \\ (\scriptsize{$\pm$}0.0273)} & \makecell[c]{0.1135 \\ (\scriptsize{$\pm$}0.0268)} & \makecell[c]{0.1964 \\ (\scriptsize{$\pm$}0.0344)} & \makecell[c]{0.1131 \\ (\scriptsize{$\pm$}0.0691)} \\ 
        \midrule

        GUETIICI & 6 & \makecell[c]{0.8660 \\ (\scriptsize{$\pm$}0.1160)} & \makecell[c]{0.7779 \\ (\scriptsize{$\pm$}0.1425)} & \makecell[c]{0.9352 \\ (\scriptsize{$\pm$}0.1078)} & \makecell[c]{0.0043 \\ (\scriptsize{$\pm$}0.0020)} & \makecell[c]{0.0022 \\ (\scriptsize{$\pm$}0.0010)} & \makecell[c]{0.0232 \\ (\scriptsize{$\pm$}0.0096)} & \makecell[c]{0.0000 \\ (\scriptsize{$\pm$}0.0000)} \\ 
        \midrule

        junqiangmler & 7 & \makecell[c]{0.3686 \\ (\scriptsize{$\pm$}0.4548)} & \makecell[c]{0.3419 \\ (\scriptsize{$\pm$}0.4220)} & \makecell[c]{0.3981 \\ (\scriptsize{$\pm$}0.4911)} & \makecell[c]{0.0899 \\ (\scriptsize{$\pm$}0.1149)} & \makecell[c]{0.0766 \\ (\scriptsize{$\pm$}0.0984)} & \makecell[c]{0.1075 \\ (\scriptsize{$\pm$}0.1370)} & \makecell[c]{0.0875 \\ (\scriptsize{$\pm$}0.1176)} \\

        \bottomrule
    \end{tabular}
    \end{adjustbox}
\end{table*}


\subsection{Evaluation and Key Findings}
\subsubsection{Evaluation and Key Findings in CBCT-IOS Registration Track }
Performance in the CBCT–IOS registration track was evaluated using a geometric error framework that quantifies the accuracy of the estimated rigid transformation aligning an intraoral scan (IOS) to a cone-beam CT (CBCT) volume. The primary metrics were mean translation error (in millimeters) and mean rotation error (in degrees), computed over corresponding anatomical landmarks in the test set. These metrics directly reflect the clinical usability of the registered models, where sub-millimeter and sub-degree precision is often required for surgical planning.
The results reveal a pronounced performance gap between competing approaches. The top-performing team, DiceMed, achieved a mean translation error of 46.47 mm and a mean rotation error of 165.30°. Their success stems from a PointNetLK-style iterative registration architecture combined with a two-stage semi-supervised training protocol: initial supervised training on labeled pairs followed by pseudo-label generation and filtering on unlabeled data, enabling effective use of scarce annotations.
In contrast, the second-place team, jichangkai, reported significantly higher errors—161.08 mm in translation and 164.58° in rotation—despite building upon the same official baseline framework. This discrepancy highlights the sensitivity of registration performance to architectural and training nuances, and underscores the effectiveness of DiceMed’s iterative refinement and semi-supervised strategy in handling real-world challenges such as partial visibility and sparse anatomical overlap.
Notably, both submissions prioritized robustness over speed, as the challenge emphasized geometric fidelity rather than real-time inference. Future work may explore the trade-off between accuracy and efficiency, but in this edition, minimizing spatial misalignment remained the paramount objective.

\subsection{Quantitative Performance}
\subsubsection{Quantitative Performance in CBCT-IOS Registration Track }
The comprehensive performance of the top teams in the CBCT-IOS Registration Track, as detailed in Table 2, highlights the effectiveness of advanced registration frameworks for this challenging task. The most important finding is the significant performance gap between the leading teams, with the top-ranked team achieving substantially lower registration errors.

 The detailed quadrant-level analysis in Table \ref{table:results_CBCT-IOS}On the test set, the best-performing team, Dice Med (rank 1), achieved an average translation error of 46.47 mm and an average rotation error of 165.30 degrees. In contrast, the second-ranked team, jichangkai (rank 2), had much higher errors, with an average translation error of 161.08 mm and an average rotation error of 164.576 degrees. This stark contrast emphasizes the critical impact of the selected algorithmic approach on the final registration accuracy. The leading methods employed distinct strategies tailored for point cloud registration. Their method utilizes a PointNetLK-style iterative registration module, which extracts features from both IOS and CBCT point clouds and iteratively predicts incremental 6-DoF transformations to refine the alignment. Crucially, they employed a two-stage semi-supervised training protocol: first training on labeled data, then generating and filtering pseudo-labels from unlabeled data to fine-tune the model, effectively mitigating the scarcity of annotated clinical data.
The second-ranked team, jichangkai, implemented a semi-supervised framework that extends the official STSR Task-2 Baseline, indicating a shared foundation but likely differing in specific architectural choices or training details, which resulted in their markedly higher error rates.
The high variance levels observed in the table (e.g., ±200.40 for Dice Med's translation error) may be attributed to the presence of outlier cases with significant registration failure and the non-percentage nature of the measurement method, where even a single large error can drastically affect the mean.

\begin{table*}[ht]
    \centering
    \caption{Image-level performance for top 2 teams in the CBCT-IOS Registration Track.The bottom table details the Instance DSC scores broken down by mean translation error and mean rotation error. The high variance level observed may be attributed to the presence of cases with significant error and non-percentage measurement method.}
    \label{table:results_CBCT-IOS}
    \resizebox{\textwidth}{!}{
    \begin{tabular}{lc cc}
        \toprule
        \multicolumn{2}{c}{\textbf{Team}} & \multicolumn{2}{c}{\textbf{Image-level}} \\
        \cmidrule{1-2} \cmidrule{3-4}
        Name & Rank & Mean Translation Error mm $\downarrow$ & Mean Rotation Error deg $\downarrow$ \\
        \midrule
        Dice Med & 1 & 46.47 {\scriptsize($\pm$200.40)} & 165.30 {\scriptsize($\pm$83.87)} \\
        jichangkai & 2 & 161.08 {\scriptsize($\pm$12986.90)} & 164.57 {\scriptsize($\pm$106.48)} \\
                \midrule
        \multicolumn{2}{c}{baseline} & 217.82 {\scriptsize($\pm$13594.51)} & 156.98 {\scriptsize($\pm$142.37)} \\
        \bottomrule
    \end{tabular}}
\end{table*}

\subsection{Analysis of Methodological Impact}
A deeper analysis reveals a clear connection between architectural choices, SSL strategies, and performance outcomes. The exceptional boundary accuracy (NSD) of Chen Liangyu can be attributed to its innovative CMambaBlock and TAAM, which enhanced the model's ability to integrate long-range context and cross-dimensional features. The success of Zhi Qin, the overall winner, stemmed from its systematic, multi-stage SSL pipeline, which enabled robust feature learning from the unlabeled data. The top IA score from DiceMed highlights that even a straightforward pseudo-labeling strategy can be supremely effective for the instance identification aspect of the task.
Analysis of failure cases across the top solutions revealed systematic challenges related to fine anatomical structures. The paper from DiceMed noted a primary weakness in segmenting the apical third of the tooth root, where low contrast-to-noise ratio (CNR) and partial volume effects led to incomplete masks. Similarly, the Zhi Qin solution struggled with predicting the precise thickness and length of the pulp canal and was susceptible to generating false positives in Limited Field of View (LFOV) scans.

\subsection{Qualitative Observations}
Qualitative assessment of the segmentation results aligns with the quantitative metrics, confirming that leading methods could reconstruct complex dental anatomy while also revealing their shared limitations. Visualizations from Ji Changkai and DiceMed showed accurate and anatomically coherent segmentations in standard cases. However, common failure modes emerged in challenging scenarios. The incomplete segmentation of root apices was a frequently observed issue, resulting in fragmented masks and anatomically incorrect 3D reconstructions. Furthermore, the imprecise delineation of fine internal structures like the pulp canal and the generation of false positives in non-standard LFOV scans indicate that while SSL has greatly advanced performance, systematic challenges remain in handling low-contrast micro-structures and improving generalization across varied imaging conditions.

\begin{figure*}[!ht]
    \centering
    \includegraphics[width=\linewidth]{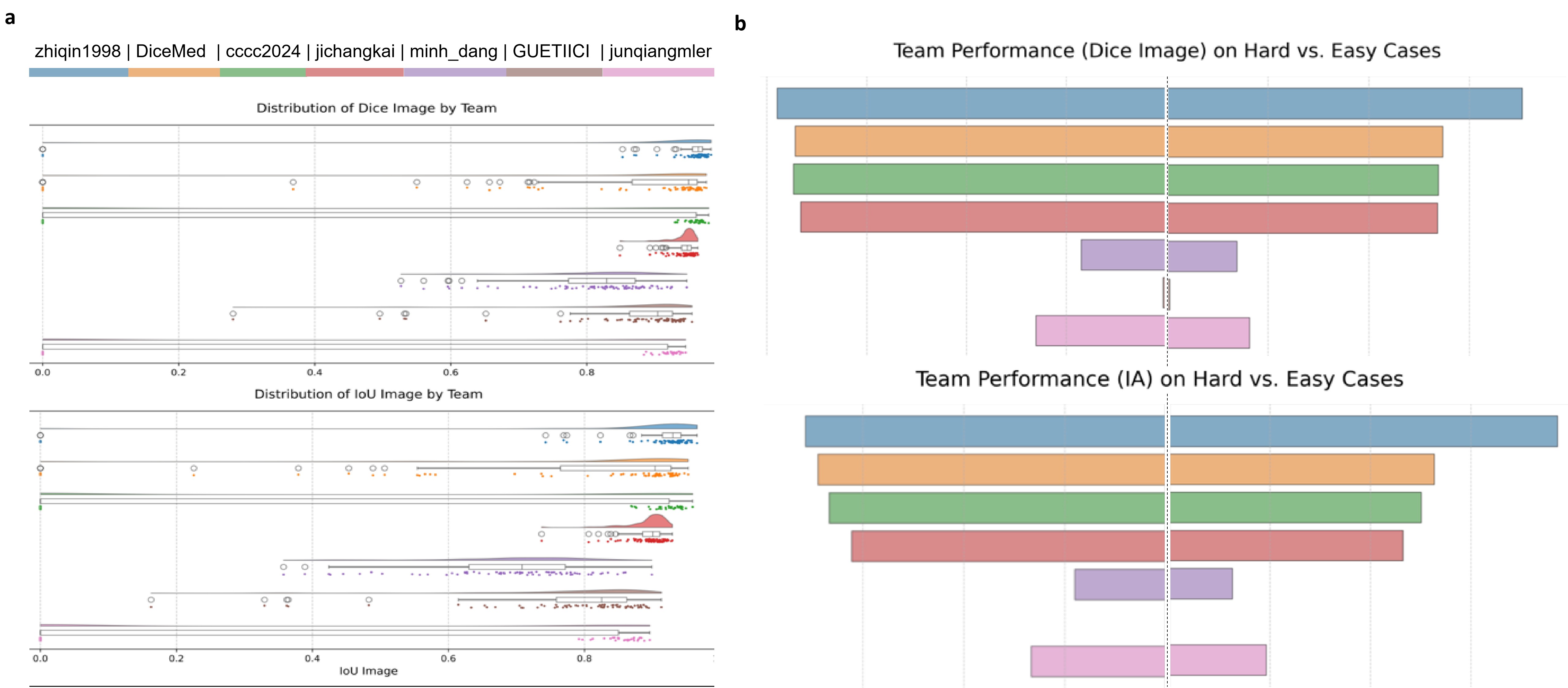}
    \caption{Performance analysis based on metric stability and case difficulty. (a) Statistical distribution of segmentation metrics (Dice and IoU) visualizing the performance variance for the participating teams. (b) Comparative analysis of model robustness on hard versus easy cases for Dice Image and IA metrics.}
    \label{fig:rainplot_adultandcild}
\end{figure*}

\begin{table*}[t!]
    \centering
    \caption{Quantitative comparison of segmentation performance for \textbf{Macroscopic Structure}. Values are presented as mean $\pm$ standard deviation.}
    \label{table:results_macro}
    \renewcommand{\arraystretch}{1.4}
    \begin{adjustbox}{width=\textwidth}
        \begin{tabular}{rccccccc}
        \hline
        \textbf{Rank} & \textbf{Team} & \textbf{Dice\_Image} & \textbf{Dice\_Instance} & \textbf{IoU\_Image} & \textbf{IoU\_Instance} & \textbf{NSD\_Image} & \textbf{NSD\_Instance} \\
        \hline
        1  & zhiqin1998   & 0.8701 $\pm$ 0.0762 & 0.8701 $\pm$ 0.0762 & 0.7889 $\pm$ 0.1103 & 0.7889 $\pm$ 0.1103 & 0.9719 $\pm$ 0.0192 & 0.9719 $\pm$ 0.0192 \\
        2  & DiceMed      & 0.8217 $\pm$ 0.0629 & 0.8217 $\pm$ 0.0629 & 0.7196 $\pm$ 0.0908 & 0.7196 $\pm$ 0.0908 & 0.9244 $\pm$ 0.0157 & 0.9244 $\pm$ 0.0157 \\
        4  & cccc2024     & 0.8693 $\pm$ 0.0788 & 0.8693 $\pm$ 0.0788 & 0.7850 $\pm$ 0.1165 & 0.7850 $\pm$ 0.1165 & 0.9797 $\pm$ 0.0173 & 0.9797 $\pm$ 0.0173 \\
        6  & jichangkai   & 0.8431 $\pm$ 0.0831 & 0.8431 $\pm$ 0.0831 & 0.7447 $\pm$ 0.1186 & 0.7447 $\pm$ 0.1186 & 0.9726 $\pm$ 0.0179 & 0.9726 $\pm$ 0.0179 \\
        8  & minh\_dang   & 0.6793 $\pm$ 0.0758 & 0.6793 $\pm$ 0.0758 & 0.5284 $\pm$ 0.0883 & 0.5284 $\pm$ 0.0883 & 0.8284 $\pm$ 0.0113 & 0.8284 $\pm$ 0.0113 \\
        9  & GUETIICI     & 0.0164 $\pm$ 0.0164 & 0.0164 $\pm$ 0.0164 & 0.0084 $\pm$ 0.0084 & 0.0084 $\pm$ 0.0084 & 0.0450 $\pm$ 0.0450 & 0.0450 $\pm$ 0.0450 \\
        11 & junqiangmler & 0.7787 $\pm$ 0.1198 & 0.7787 $\pm$ 0.1198 & 0.6550 $\pm$ 0.1614 & 0.6550 $\pm$ 0.1614 & 0.9512 $\pm$ 0.0408 & 0.9512 $\pm$ 0.0408 \\
        12 & baseline     & 0.8025 $\pm$ 0.1285 & 0.8025 $\pm$ 0.1285 & 0.6925 $\pm$ 0.1787 & 0.6925 $\pm$ 0.1787 & 0.9773 $\pm$ 0.0192 & 0.9773 $\pm$ 0.0192 \\
        \hline
        \end{tabular}
    \end{adjustbox}
\end{table*}

\begin{table*}[t!]
    \centering
    \caption{Quantitative comparison of segmentation performance for \textbf{Root Canal System}. Values are presented as mean $\pm$ standard deviation.}
    \label{table:results_root}
    \renewcommand{\arraystretch}{1.4}
    \begin{adjustbox}{width=\textwidth}
        \begin{tabular}{rccccccc}
        \hline
        \textbf{Rank} & \textbf{Team} & \textbf{Dice\_Image} & \textbf{Dice\_Instance} & \textbf{IoU\_Image} & \textbf{IoU\_Instance} & \textbf{NSD\_Image} & \textbf{NSD\_Instance} \\
        \hline
        1  & zhiqin1998   & 0.6409 $\pm$ 0.0455 & 0.6470 $\pm$ 0.0389 & 0.4945 $\pm$ 0.0462 & 0.4992 $\pm$ 0.0415 & 0.9339 $\pm$ 0.0352 & 0.9430 $\pm$ 0.0173 \\
        2  & DiceMed      & 0.6199 $\pm$ 0.0488 & 0.6230 $\pm$ 0.0460 & 0.4634 $\pm$ 0.0498 & 0.4656 $\pm$ 0.0478 & 0.9329 $\pm$ 0.0236 & 0.9376 $\pm$ 0.0139 \\
        4  & cccc2024     & 0.6137 $\pm$ 0.0469 & 0.6202 $\pm$ 0.0470 & 0.4577 $\pm$ 0.0484 & 0.4625 $\pm$ 0.0487 & 0.9335 $\pm$ 0.0336 & 0.9428 $\pm$ 0.0149 \\
        6  & jichangkai   & 0.5886 $\pm$ 0.0436 & 0.5927 $\pm$ 0.0407 & 0.4318 $\pm$ 0.0429 & 0.4348 $\pm$ 0.0410 & 0.9273 $\pm$ 0.0307 & 0.9338 $\pm$ 0.0177 \\
        8  & minh\_dang   & 0.0017 $\pm$ 0.0015 & 0.0017 $\pm$ 0.0015 & 0.0008 $\pm$ 0.0008 & 0.0008 $\pm$ 0.0008 & 0.0284 $\pm$ 0.0227 & 0.0290 $\pm$ 0.0236 \\
        9  & GUETIICI     & 0.0013 $\pm$ 0.0030 & 0.0013 $\pm$ 0.0030 & 0.0007 $\pm$ 0.0015 & 0.0007 $\pm$ 0.0015 & 0.0216 $\pm$ 0.0434 & 0.0216 $\pm$ 0.0434 \\
        11 & junqiangmler & 0.0000 $\pm$ 0.0000 & 0.0000 $\pm$ 0.0000 & 0.0000 $\pm$ 0.0000 & 0.0000 $\pm$ 0.0000 & 0.0000 $\pm$ 0.0000 & 0.0000 $\pm$ 0.0000 \\
        12 & baseline     & 0.0761 $\pm$ 0.1703 & 0.0761 $\pm$ 0.1703 & 0.0509 $\pm$ 0.1139 & 0.0509 $\pm$ 0.1139 & 0.1502 $\pm$ 0.3359 & 0.1502 $\pm$ 0.3359 \\
        \hline
        \end{tabular}
    \end{adjustbox}
\end{table*}

\begin{table*}[t!]
    \centering
    \caption{Quantitative comparison of segmentation performance for \textbf{Special Condition}. Values are presented as mean $\pm$ standard deviation.}
    \label{table:results_special}
    \renewcommand{\arraystretch}{1.4}
    \begin{adjustbox}{width=\textwidth}
        \begin{tabular}{rccccccc}
        \hline
        \textbf{Rank} & \textbf{Team} & \textbf{Dice\_Image} & \textbf{Dice\_Instance} & \textbf{IoU\_Image} & \textbf{IoU\_Instance} & \textbf{NSD\_Image} & \textbf{NSD\_Instance} \\
        \hline
        1  & zhiqin1998   & 0.9025 $\pm$ 0.0000 & 0.9527 $\pm$ 0.0000 & 0.8634 $\pm$ 0.0000 & 0.9114 $\pm$ 0.0000 & 0.9397 $\pm$ 0.0000 & 0.9919 $\pm$ 0.0000 \\
        2  & DiceMed      & 0.7388 $\pm$ 0.0000 & 0.8921 $\pm$ 0.0000 & 0.6759 $\pm$ 0.0000 & 0.8162 $\pm$ 0.0000 & 0.7425 $\pm$ 0.0000 & 0.8965 $\pm$ 0.0000 \\
        4  & cccc2024     & 0.8660 $\pm$ 0.0000 & 0.9353 $\pm$ 0.0000 & 0.8167 $\pm$ 0.0000 & 0.8821 $\pm$ 0.0000 & 0.9120 $\pm$ 0.0000 & 0.9849 $\pm$ 0.0000 \\
        6  & jichangkai   & 0.8293 $\pm$ 0.0000 & 0.8875 $\pm$ 0.0000 & 0.7588 $\pm$ 0.0000 & 0.8121 $\pm$ 0.0000 & 0.8965 $\pm$ 0.0000 & 0.9595 $\pm$ 0.0000 \\
        8  & minh\_dang   & 0.0000 $\pm$ 0.0000 & 0.0000 $\pm$ 0.0000 & 0.0000 $\pm$ 0.0000 & 0.0000 $\pm$ 0.0000 & 0.0000 $\pm$ 0.0000 & 0.0000 $\pm$ 0.0000 \\
        9  & GUETIICI     & 0.0000 $\pm$ 0.0000 & 0.0000 $\pm$ 0.0000 & 0.0000 $\pm$ 0.0000 & 0.0000 $\pm$ 0.0000 & 0.0000 $\pm$ 0.0000 & 0.0000 $\pm$ 0.0000 \\
        11 & junqiangmler & 0.6390 $\pm$ 0.0000 & 0.7912 $\pm$ 0.0000 & 0.5637 $\pm$ 0.0000 & 0.6979 $\pm$ 0.0000 & 0.7154 $\pm$ 0.0000 & 0.8858 $\pm$ 0.0000 \\
        12 & baseline     & 0.9113 $\pm$ 0.0000 & 0.9273 $\pm$ 0.0000 & 0.8533 $\pm$ 0.0000 & 0.8683 $\pm$ 0.0000 & 0.9677 $\pm$ 0.0000 & 0.9847 $\pm$ 0.0000 \\
        \hline
        \end{tabular}
    \end{adjustbox}
\end{table*}


\subsubsection{Quantitative Evaluation in 3D CBCT Track}

The 3D CBCT segmentation track posed a significant challenge, characterized by high-dimensional data and a high prevalence of imaging artifacts. This complexity was reflected in the final submissions, with only five teams successfully completing the task. The results, however, provide a powerful demonstration of semi-supervised learning's efficacy in this demanding volumetric domain.

The most striking outcome of the 3D track is the dramatic performance improvement conferred by SSL. A fully-supervised 3D nnU-Net baseline, trained only on the 30 labeled CBCT scans, achieved a modest Instance DSC of 30.80\%. In contrast, the winning SSL method from team ChohoTech achieved an Instance DSC of 92.15\%. This represents an absolute gain of over 61 percentage points (a relative improvement of 197\%), decisively validating the use of SSL for volumetric dental instance segmentation and proving its ability to overcome extreme data scarcity.

As detailed in Table \ref{table:results_3d} and visualized in Fig. \ref{fig:rainplot_3DCBCT}(a), team ChohoTech's approach, which adapted a YOLO-based pipeline to 3D, was not only the most accurate but also the most stable, evidenced by its low standard deviations across all metrics. This stability contrasts sharply with the performance of several other teams, including the second-place winner Houwentai, which exhibited very high variance (e.g., a standard deviation of 33.82\% for Instance DSC). This suggests that their SSL strategies, while effective on some cases, struggled to generalize across the full diversity of the unlabeled test set. This instability may indicate that their pseudo-labeling or consistency schemes were sensitive to domain shifts introduced by clinical factors like metal artifacts, causing degraded performance on unlabeled data that differed significantly from the small labeled set.

The quadrant-level analysis in Table \ref{table:results_3d_quadrant} and Fig. \ref{fig:rainplot_3DCBCT}(b) further reinforces the robustness of the top methods, which maintained consistently high performance across all four anatomical quadrants. The clinical relevance of these results is underscored by the excellent Normalized Surface Distance (NSD) scores achieved by the top teams. A high NSD score indicates the generation of segmentations with highly accurate surface boundaries, which is a critical prerequisite for clinical applications such as the digital design and fabrication of precise orthodontic appliances and surgical guides.

\subsubsection{Comparative Analysis between 2D and 3D Tracks}

A cross-task comparison reveals differences in performance, model complexity, and computational cost between the 2D PXI and 3D CBCT tracks, highlighting the distinct challenges of each modality. Four teams: ChohoTech, Jichangkai, Junqiangmler, and Madongdong competed in both tracks, allowing for a direct comparison of how their strategies adapted across dimensions.

The most striking difference lies in the potential for segmentation accuracy. As shown in Table \ref{table:results_2d} and Table \ref{table:results_3d}, the winning team ChohoTech achieved a remarkable Instance DSC of 92.15\% in the 3D track, substantially higher than their already impressive 83.59\% in the 2D track. This suggests that the additional spatial information in 3D data is highly effective at resolving ambiguities inherent in 2D projections, most notably the superposition of tooth structures that complicates boundary delineation, leading to more accurate and robust instance segmentation once a model can effectively process volumetric data. However, capitalizing on this 3D advantage proved non-trivial. Other teams like Jichangkai and Junqiangmler saw a decrease in their average scores when moving from 2D to 3D, underscoring the challenge of successfully adapting SSL methods to a higher-dimensional space.

This difficulty is intrinsically linked to the algorithmic efficiency, where the computational burden of the 3D track was an order of magnitude greater. For ChohoTech, the average runtime (RT) increased from 13.29 seconds per 2D image to 60.76 seconds per 3D volume. More dramatically, their integrated GPU memory usage (AUC\_GPU) surged from approximately 7,341 GB$\cdot$s to 233,660 GB$\cdot$s. This massive increase in resource requirements explains why fewer teams were able to complete the 3D task and emphasizes a critical trade-off for clinical deployment: while 3D models can yield superior accuracy, their high computational cost may limit their practical application. The semi-supervised learning paradigm itself is also more demanding in 3D, as generating and refining pseudo-labels for volumetric data requires significantly more memory and computational power, a factor that contributes to the performance instability observed in several 3D submissions.

\begin{figure*}[!ht]
    \centering
    \includegraphics[width=0.8\linewidth]{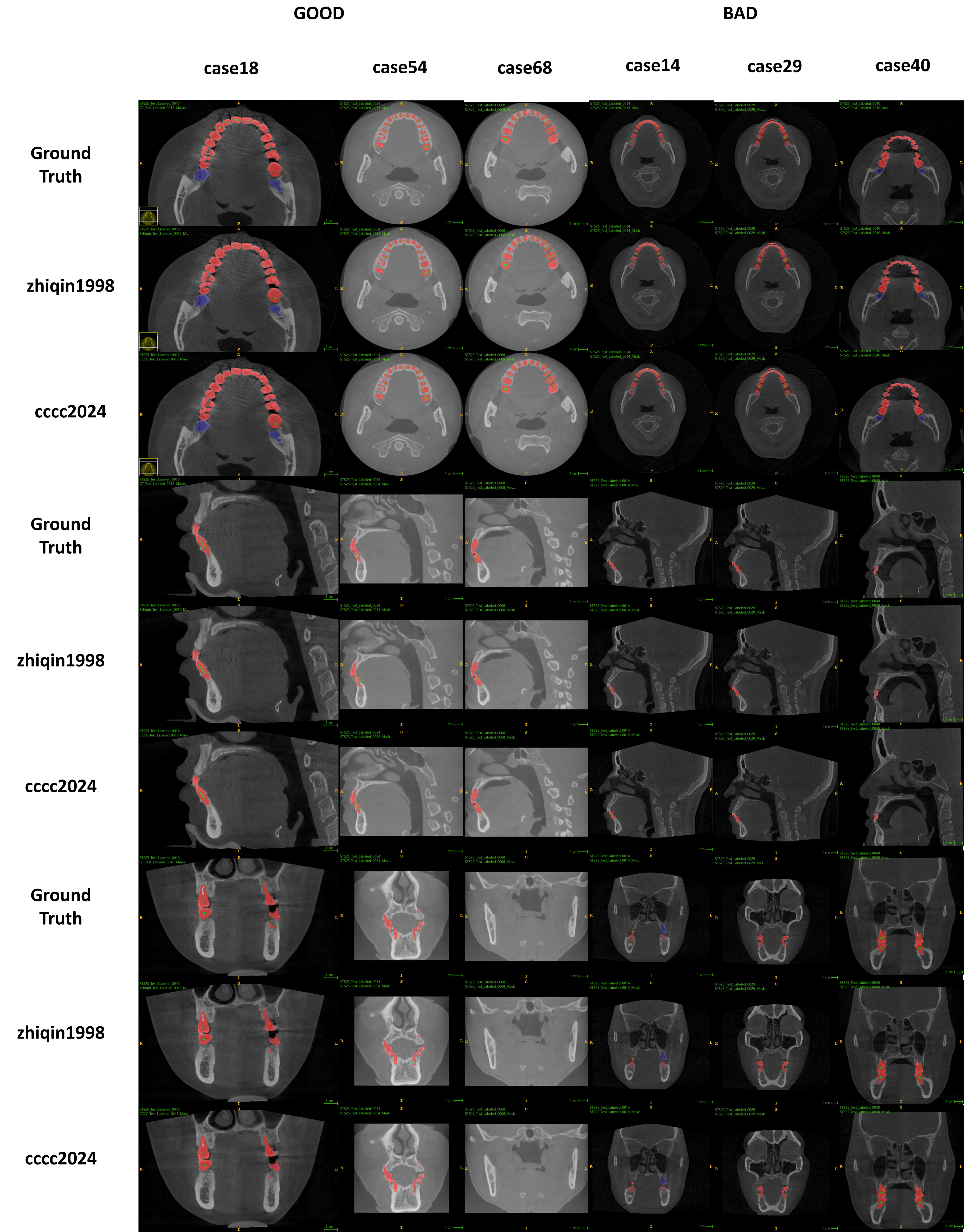}
    \caption{Visual comparison of 3D CBCT segmentation results.}
    \label{fig:3DCBCT_3dview}
\end{figure*}

\begin{figure*}[!ht]
    \centering
    \includegraphics[width=\linewidth]{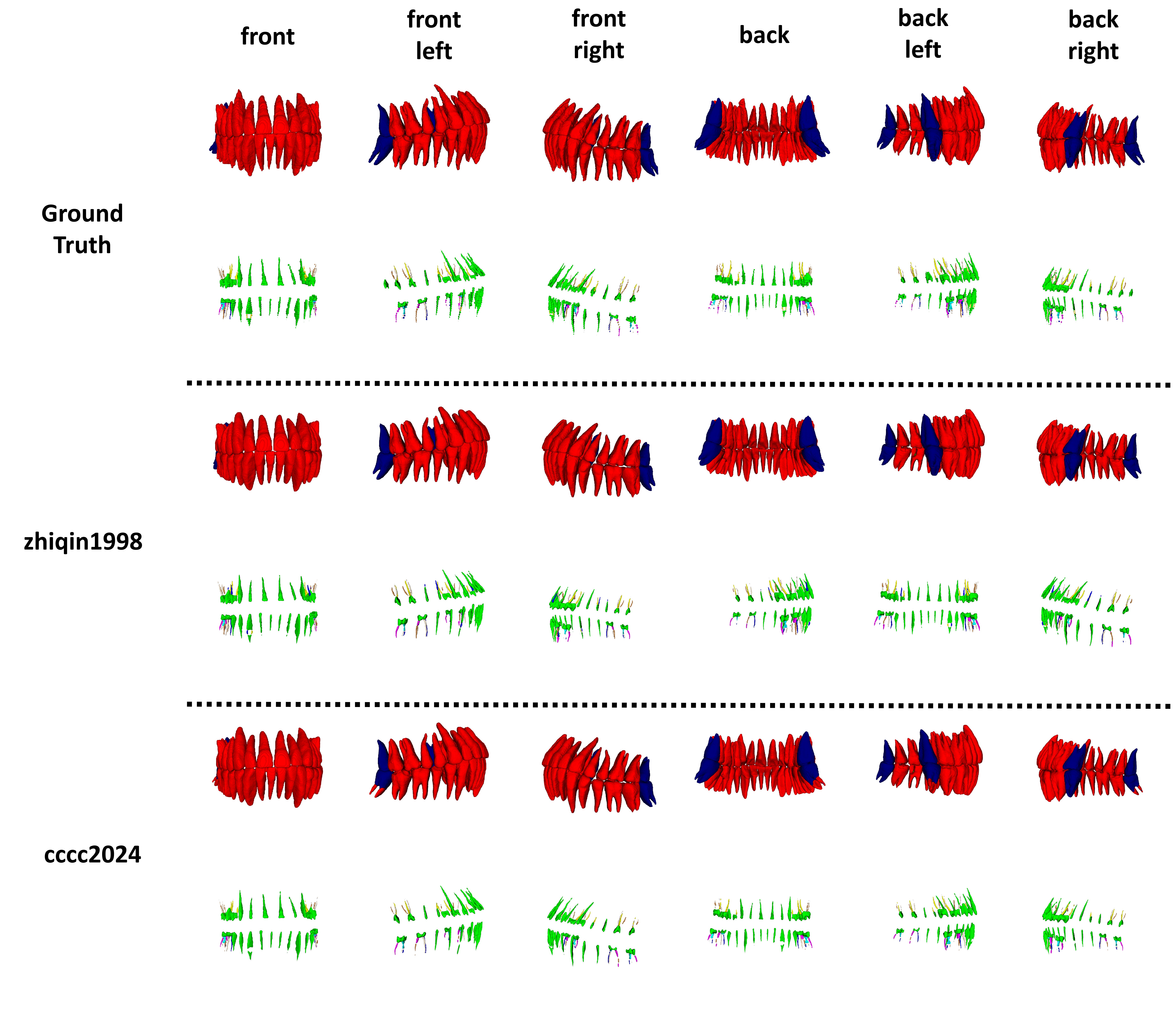}
    \caption{Visual comparison of 3D CBCT segmentation results. Multi-perspective views of the ground truth (a) are contrasted with outputs from top teams Houwentai and Jichangkai (b, c). Specific failure modes, including lack of segmentation (L), over-segmentation (O), and recognition errors (R) are indicated. A 2D slice comparison is provided in (d) for boundary detail.}
    \label{fig:3DCBCT_3dview}
\end{figure*}

\subsection{Qualitative Analysis of Segmentation Failures and Successes}

To complement the quantitative metrics, a qualitative analysis of the segmentation results provides crucial insights into the specific clinical and anatomical challenges that current semi-supervised methods face.

\subsubsection{Analysis of 2D PXI Segmentation}
Visual inspection of adult cases in Fig. \ref{fig:2DPXI_adult} reveals that while top-performing methods were proficient in segmenting well-defined teeth, two primary failure modes emerged in challenging regions. First, the erroneous merging of overlapping teeth (circled in red) was a common issue. In 2D panoramic projections, the boundaries between crowded or superimposed teeth become ambiguous. This suggests that SSL models relying heavily on local pixel context can struggle to enforce instance separation without a strong prior, a problem that detection-first architectures are designed to mitigate. Second, many methods produced inaccurate segmentations of tooth apices (circled in blue). The apex is a small, low-contrast structure of high clinical importance for diagnosing periapical lesions. Its poor segmentation likely indicates that standard region-based loss functions (e.g., Dice) are dominated by the larger tooth crown, thus failing to penalize errors on these small but critical anatomical targets.

These challenges were significantly amplified in the pediatric cases shown in Fig. \ref{fig:2DPXI_child}. Pediatric dentition, with its mix of smaller deciduous teeth and developing permanent teeth, creates a dense and highly irregular environment. The magnified views demonstrate that most methods struggled in these cluttered regions, often failing to delineate individual small teeth. This highlights a key difficulty for SSL: if the limited labeled data does not sufficiently represent the vast anatomical variability of pediatric development, the model may generate noisy or incomplete pseudo-labels for unlabeled pediatric cases, hindering its ability to learn robust features for these challenging objects.


    
\subsubsection{Qualitative Analysis of 3D Tracks at zoomed in view}

For the 3D track, Fig. \ref{fig:3DCBCT_3dview} illustrates typical error modes in volumetric segmentation. While leading methods successfully reconstructed the 3D morphology of most teeth, three primary failure patterns emerged: under-segmentation (e.g., merging adjacent teeth, labeled 'L'), over-segmentation (e.g., erroneous extensions into surrounding bone, labeled 'O'), and recognition errors (e.g., correct segmentation but incorrect tooth identification, labeled 'R'). These qualitative observations corroborate the quantitative findings, underscoring that accurate instance separation and identification remain critical challenges in 3D CBCT segmentation.





\begin{figure}[!ht]
    \centering
    \includegraphics[width=\linewidth]{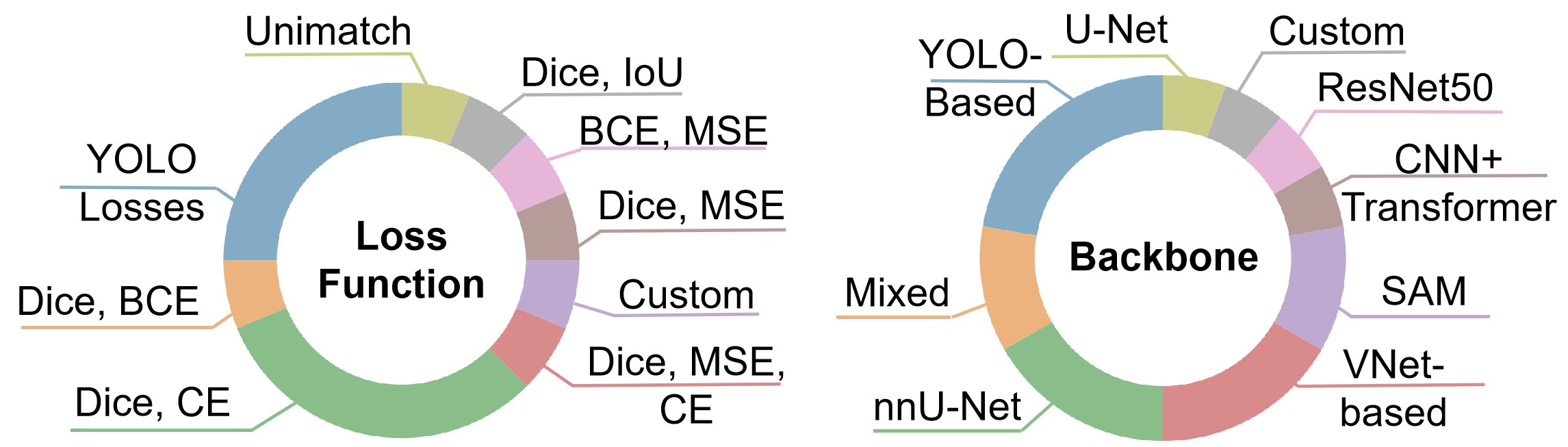}
    \caption{Distribution of loss functions employed and backbone architectures by participants in the 2D and 3D challenge tracks. (Left) Analysis of loss function usage indicates a strong preference for composite loss functions over single-objective training. The combination of a region-based loss (Dice) and a pixel-wise loss (Cross-Entropy) was the most common strategy. (Right) Breakdown of backbone model choices, revealing that nnU-Net and V-Net were the most prevalent architectures. A notable trend was the integration of the Segment Anything Model (SAM) as a foundational component. }
    \label{fig:backboneandloss}
\end{figure}

\subsection{Methodological Insights from Leading Teams in CBCT-IOS registration}
An examination of the public implementations and descriptions released by leading Task-2 participants shows a clear convergence toward a registration-centric design philosophy: compact, point-cloud oriented feature extractors, iterative pose refinement, and pragmatic semi-supervised training pipelines. Two representative public repositories — one implementing a PointNetLK-style iterative registration module and another providing an extensible baseline incorporating differentiable geometric solvers — exemplify these choices and the practical engineering trade-offs made for robust CBCT-IOS alignment.

\subsubsection{Network Architectures}
Top submissions favor lightweight point-set feature encoders rather than heavy volumetric backbones. A common pattern is a stack of 1D convolution layers (Conv1D - BN - ReLU) to extract per-point descriptors, followed by a global pooling operation to produce compact scene descriptors. The feature difference between IOS and CBCT descriptors is then processed to estimate the pose. While some teams employed an update network to predict incremental 6-DoF transforms similar to PointNetLK, others utilized a differentiable Singular Value Decomposition (SVD) head to explicitly solve for the optimal rigid transformation. This design handles partial overlaps and variable point densities typical in dental scans while keeping inference efficient.

\subsubsection{Semi-supervised Techniques}
The semi-supervised schemes implemented in the repositories follow a two-stage protocol. First, models are trained on the available labeled scan pairs to establish a stable initialization. Second, the trained model generates pseudo-labels on unlabeled pairs; low-confidence predictions are filtered and only reliable pseudo-labels are mixed with true labels for fine-tuning. This staged approach reduces confirmation bias from noisy pseudo-labels and improves generalization. Practical engineering details used in the released code include confidence thresholding, iterative re-estimation of pseudo-labels, and separate learning-rate schedules for the pseudo-label fine-tuning phase.

\subsubsection{Loss Function and optimization targets}
Reported implementations combine geometric and feature-level objectives. Typical terms include a descriptor-level matching loss (to align learned features across modalities), a transform regression loss (L2 on predicted incremental pose parameters), and geometric alignment losses (e.g., point-to-point Chamfer distance). In practice, teams found that balancing descriptor matching with a final geometric distance improves both convergence speed and final registration accuracy.

\subsubsection{Modular, multi-stage pipelines and practical components}
Rather than one monolithic network, the released solutions favor modular pipelines: robust preprocessing (e.g., HU thresholding for CBCT), a compact feature encoder, and a post-refinement stage. This modularity simplifies debugging and allows component-wise ablation. The public code also emphasizes reproducibility: clear dataset download instructions, training scripts, and inference wrappers for Dockerized evaluation.

\subsubsection{Performance across anatomical regions and demographics}
The implementation details and semi-supervised training choices above directly influence robustness. When evaluated on full-arch datasets, methods employing iterative refinement with cautious pseudo-label filtering show stable alignment across different quadrants of the dentition. Public inference scripts facilitate per-quadrant analyses, enabling teams to stratify failure modes (e.g., posterior molars with heavy restorations). Overall, the released baselines demonstrate that a well-tuned, semi-supervised point-cloud registration pipeline can deliver consistent performance across arch regions, effectively minimizing Mean Rotation and Translation Errors (MRE/MTE) even in cases of partial IOS coverage.


\section{Discussion}

\subsection{Main findings}
Dental image segmentation presents unique challenges that differentiate it from other medical imaging domains. The STS 2024 Challenge underscored several critical difficulties inherent to dental datasets: 1) Limited Segmentation Accuracy of Previous Methods; 2) Inadequate Integration of Segmentation and Tooth Position Recognition; 3) High Proportion of Unannotated Data. Traditional methods rely heavily on fully supervised learning paradigms, which demand extensively annotated datasets. However, annotating dental images is labor-intensive and requires specialized expertise, leading to a scarcity of high-quality labeled data. Consequently, previous approaches exhibited limited precision in accurately delineating tooth boundaries, especially in cases involving overlapping structures or varying image qualities. The complexity of dental anatomy, with its intricate and closely packed tooth structures, further exacerbates segmentation inaccuracies. Beyond mere segmentation, accurate tooth position identification is crucial for comprehensive dental analysis and clinical decision-making. Prior methodologies typically treated segmentation and tooth position recognition as separate tasks, leading to suboptimal performance in integrated scenarios. This discrimination fails to leverage the inter-dependencies between segmentation and positional data, resulting in inconsistencies and reduced overall accuracy. The inability to concurrently address segmentation and tooth position recognition limits the utility of automated systems in practical dental applications. The STS 2024 Challenge introduced several innovative elements in dataset construction and challenge design to address the aforementioned challenges. First, to mitigate the scarcity of annotated data, we meticulously curated a dataset encompassing 2D panoramic X-ray images and 3D CBCT tooth volumes. This multiple-modal dataset enriches the dataset and facilitates the development of algorithms capable of handling different imaging modalities, encouraging the creation of more robust and generalizable models. Furthermore, including pediatric and adult dental images ensures that models can adapt to variations across age groups, enhancing their applicability in real-world clinical settings. Our challenge incorporated a multi-faceted evaluation framework that assesses participants' algorithms on several levels, ensuring a comprehensive performance evaluation: Instance-Level Evaluation, Image-Level Evaluation.

The results from the STS 2024 Challenge reveal several noteworthy insights: Top-performing teams predominantly leveraged semi-supervised learning frameworks, effectively utilizing the limited labeled data alongside the abundant unlabeled data. Techniques such as integrating pre-trained models like the Segment Anything Model (SAM), consistency regularization learning, and multi-stage architecture optimization significantly enhanced segmentation accuracy, which aligns with the broader trend in medical imaging, where semi-supervised methods are increasingly recognized for their ability to overcome data scarcity. In the 3D CBCT segmentation task, multi-stage training strategies yielded superior performance. Participants could incrementally improve segmentation precision and robustness by sequentially refining the model through multiple training phases. This approach facilitates better feature learning and mitigates the risk of overfitting, particularly in complex 3D structures. Despite the overall success of semi-supervised methods, a notable portion of participants did not employ any semi-supervised strategies. This variability reveals a critical gap between the demonstrated potential of SSL and its practical adoption, which have demonstrated clear advantages in handling unannotated data. Across the diverse methodologies, participants predominantly introduced perturbations at multiple levels to augment model training and generalization: Data-Level Perturbations: Techniques such as weak-strong augmentations and the addition of Gaussian noise were employed to create varied input data representations, fostering robustness against input variations. Model-Level Perturbations: The integration of heterogeneous network architectures (e.g., CNNs and Transformers) facilitated implicit consistency regularization, promoting diverse feature learning and reducing model-specific biases. Training Cycle Perturbations: Frameworks like the teacher-student model incorporated temporal perturbations by iteratively refining pseudo-labels, thereby enhancing feature learning and mitigating overfitting. In terms of consistency constraints, methods varied across three primary dimensions: Soft Constraints: Utilization of soft logits and features as supervisory signals encouraged smooth decision boundaries and feature consistency across different perturbations. Hard Constraints: Binary segmentation outputs were enforced through hard label assignments, ensuring definitive delineation of anatomical structures. Structural Constraints: Incorporation of edge detection operators (e.g., Sobel) imposed structural integrity on anatomical boundaries, aligning segmentation outputs with inherent anatomical features.

Building upon the insights gained from the STS 2024 Challenge, several avenues for future research and development emerge. First, efforts should be directed toward creating more extensive and diverse annotated dental datasets encompassing various anatomical variations, imaging modalities, and clinical conditions. Collaborative initiatives and data-sharing platforms can facilitate the accumulation of comprehensive datasets necessary for training robust deep-learning models. Further exploration of semi-supervised and unsupervised learning methods can enhance model performance and reduce dependency on labeled data. Techniques such as self-supervised learning, transfer learning, and active learning hold promise for improving segmentation accuracy and efficiency. Developing models that concurrently address segmentation and tooth position identification can yield more holistic and clinically relevant outputs. Multi-task learning frameworks can enhance the synergy between tasks, improving overall performance and utility. Bridging the gap between research and clinical practice requires rigorous validation of segmentation models in real-world settings. Future studies should focus on integrating automated segmentation tools into clinical workflows and assessing their impact on diagnostic accuracy, treatment planning, and patient outcomes. Enhancing the computational efficiency of semi-supervised models is crucial for their adoption in clinical environments. Research should optimize models for real-time processing without compromising segmentation accuracy, ensuring their practicality and scalability. Subsequent iterations of the STS Challenge can incorporate more diverse and complex datasets, introduce additional evaluation metrics that reflect clinical utility, and encourage the development of lightweight models suitable for deployment in various clinical settings. Such enhancements can further drive innovation and accelerate the adoption of advanced segmentation techniques in dentistry.

\begin{figure*}[!ht]
    \centering
    \includegraphics[width=\linewidth]{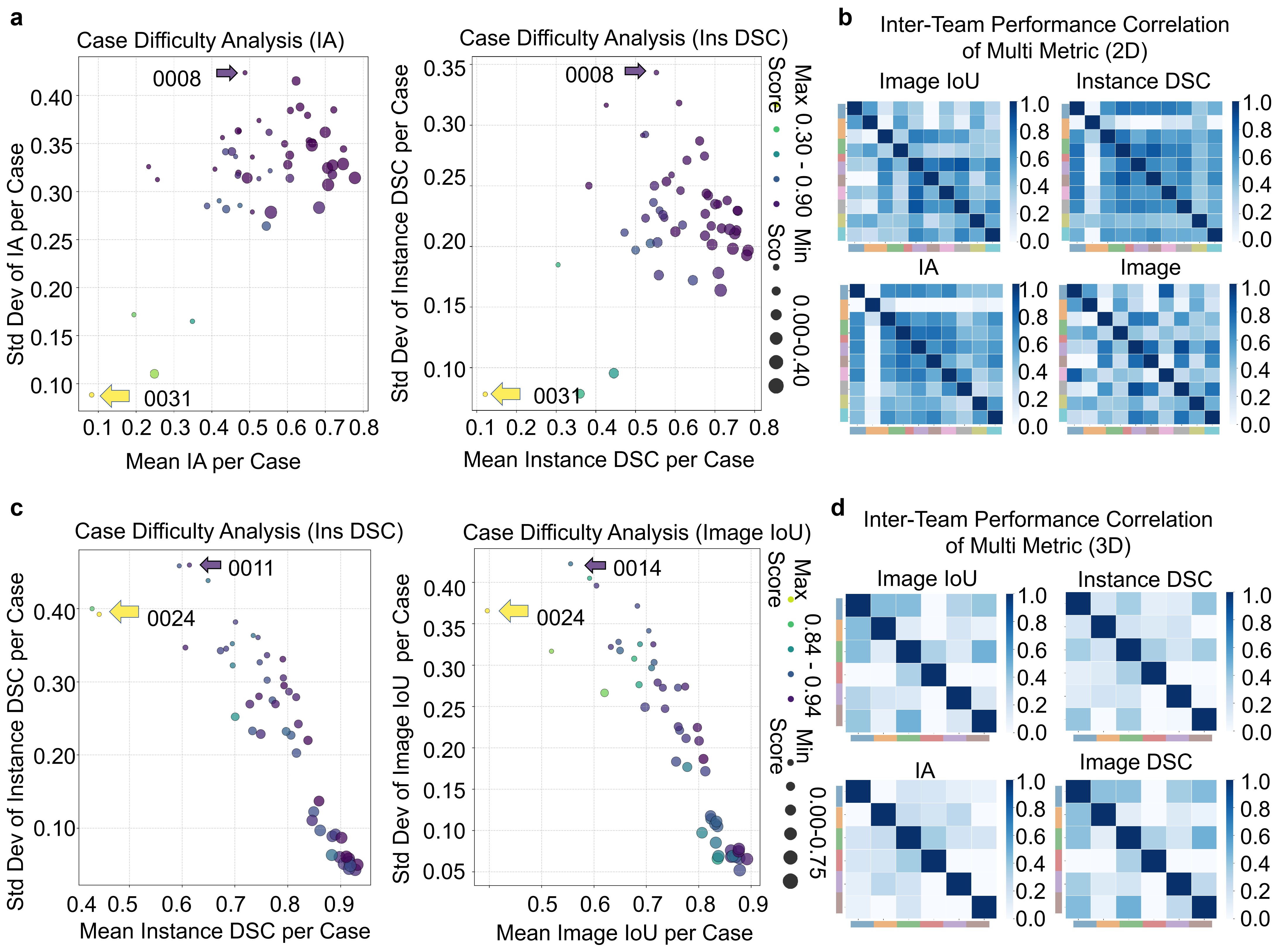}
    \caption{Case difficulty and inter-metric correlation analysis for the 2D (a, b) and 3D (c, d) tracks.
(a, c) Case Difficulty Analysis: Plots identifying the most challenging cases (0031 in 2D, 0024 in 3D). Difficulty in 2D was linked to severe tooth loss, while in 3D, it arose from small targets (e.g., apices) and inter-tooth misidentification.
(b, d) Inter-Metric Correlation: Matrices for both tracks showing strong concordance across metrics, indicating consistent team rankings.}
    \label{fig:relation_2d_01}
\end{figure*}

\subsection{Case Difficulty and Metric Correlation}

The case difficulty and metric correlation analysis in Fig.~\ref {fig:relation_2d_01} offers valuable insights into the dataset's intrinsic challenges and the robustness of our evaluation framework.

The difficulty analysis in Fig.~\ref {fig:relation_2d_01}(a, c) reveals that the primary challenges differed by modality. In the 2D PXI track, the most difficult cases were those with severe pathologies like extensive tooth loss. This is a classic problem for semi-supervised learning, as these atypical cases deviate from the primary distribution of the unlabeled data, hindering the generation of effective pseudo-labels. This suggests a need for SSL methods that are more robust to domain shift within the dataset itself. In contrast, 3D CBCT challenges were more related to fine anatomical details, such as segmenting small tooth apices and resolving ambiguity between adjacent teeth. This points to architectural limitations in current models, which may require more specialized designs to capture fine-grained features in volumetric data.

The strong positive correlation between different evaluation metrics, as shown in the matrices from ~\ref {fig:relation_2d_01}(b, d), is an interesting finding. The high concordance across metrics like Instance DSC and Instance Affinity (IA) indicates that the team rankings were stable and not dependent on a specific metric choice. This validates the robustness of our evaluation framework and suggests that the top-performing methods were genuinely superior across multiple criteria of segmentation quality.

\begin{figure*}[!ht]
    \centering
    \includegraphics[width=\linewidth]{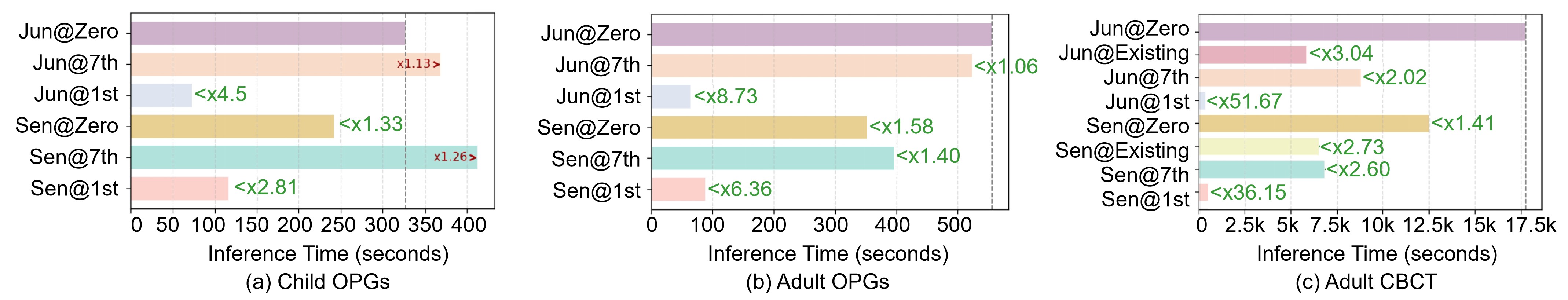}
    \caption{Bar plots of X-ray and CBCT labeling time under varied methods. The \textcolor{green}{green arrow} indicates a time reduction (acceleration) and the \textcolor{red}{red arrow} an increase (deceleration) compared to the baseline (typically the Junior@Zero). Jun: Junior dentist; Sen: Senior dentist.}
    \label{fig:acc_rates}
\end{figure*}

\subsection{Human-Machine Collaborative Annotation Study}
To quantitatively evaluate the clinical utility of semi-supervised segmentation algorithms in this challenge, we conducted a human-machine collaborative annotation study. The study involved two dentists (a junior clinician with more than two years of experience and a senior specialist with more than ten years of experience). The dentists annotated a curated test set of 30 dental scans, including 10 X-rays from child, 10 X-rays images from adult, and 10 CBCT volumes from adult. The study follows four distinct strategies:
(i) Full manual annotation (i.e, \textit{From scratch});
(ii) Correction of predictions from existing methods (only available for the CBCT);
(iii) Correction of $\text{7}^\text{th}$-ranked outputs (relatively low-quality); and
(iv) Correction of $\text{1}^\text{st}$-ranked outputs (relatively high-quality).

Annotation time per case was rigorously recorded to assess workflow efficiency, as shown in Fig.~\ref{fig:acc_rates}. This study focuses on three key dimensions: (i) the relationship between algorithmic performance (as reflected by model rankings) and time saved during clinical workflows; (ii) modality-specific challenges in 2D vs. 3D tasks, including anatomical variability across pediatric and adult populations; and (iii) how clinician expertise interacts with automated tools to optimize workflow efficiency.

\subsubsection{Dramatic Time Savings in Clinical Workflows}
The top-performing semi-supervised model (the $\text{1}^\text{st}$-ranked team) demonstrated transformative efficiency gains across all modalities. In 2D X-rays (a historically unaddressed domain), the labeling efficiency improved 4.5$\times$ for the junior dentists (i.e., from 326.10s to 72.50s per case) and 2.11$\times$ for the senior (i.e., from 244.50s to 116.10s). Critically, the $\text{1}^\text{st}$-ranked model outperformed existing CBCT methods by 94.1\% (i.e., the junior's time that decreased from 5841.80s to 343.30s), validating its superiority in handling sparse-annotation scenarios. For the CBCT volumes, the annotation time plummeted from 3.5 \~ 4.9 hours to 5.7 \~ 8.2 minutes per volume, which is clinically critical for efficient orthodontics and implant planning.

Moreover, the segmentation quality directly impacts the annotation efficiency. The relatively low-quality initial predictions may increase the cognitive load and correction time of dentists. For instance, the $\text{7}^\text{th}$-ranked model prolonged senior dentists’ child X-ray annotation by 68.5\% (i.e., increasing from 244.50s to 412.00s), indicating that subpar segmentation disrupts clinical efficiency. Conversely, the $\text{1}^\text{st}$-ranked model reduced time while improving consistency, where the standard deviations decreased across tasks. This highlights that only high-precision algorithms enable reliable human-machine collaboration, which is one of the core goal of our semi-supervised challenge.

Notably, the CBCT modality that is essential for complex orthodontic procedures, showed the greatest absolute time savings (e.g., from 4.9 hours to $\leq$0.1 hour / volume). This aligns with our challenge’s focus on underrepresented 3D data, addressing a critical barrier in dental area. For adult X-rays, the $\text{1}^\text{st}$-ranked model reduced annotation time by 88.2\% (e.g., the junior's time decreasd from 555.50s to 63.60s), outperforming the $\text{7}^\text{th}$-place model by 87.7\%. Notably, for pediatric X-rays—a domain previously lacking public datasets—the methods achieved near-adult levels of efficiency (e.g., 72.5s vs. 65.7s for the junior), proving semi-supervised methods can overcome pediatric data scarcity.

\subsubsection{Implications for Clinical Deployment}
The junior dentist benefited disproportionately from high-quality AI assistance. In pediatric X-rays, their efficiency gain with the $\text{1}^\text{st}$-ranked model (4.5$\times$) exceeded Seniors’ (2.1$\times$), narrowing the experience gap: the junior-senior time differential dropped from 81.6s (manual) to 43.6s (AI-assisted). However, Seniors exhibited lower tolerance for imperfect predictions—when using the $\text{7}^\text{th}$-place model, their pediatric X-ray annotation time surged by 68.5\%, versus only 12.9\% for Juniors. This implies that AI assistance most empowers early-career clinicians, while experts require near-perfect segmentation for workflow integration. A Senior dentist could process 96 CBCT volumes/day with $\text{1}^\text{st}$-ranked model assistance versus 2 volumes/day manually—enabling large-scale data curation for AI development. Combining Junior clinicians with state-of-the-art AI achieved 73\% of Senior-level efficiency (e.g., pediatric X-rays: 72.5s vs. Senior’s 116.1s) at lower resource cost. This validates our challenge’s design premise: semi-supervised learning bridges data scarcity in specialized domains (pediatrics, CBCT) while optimizing clinical resource utilization.

In a nutshell, our challenge’s top semi-supervised model reduced dental annotation effort by 52–98\%, demonstrating that human-AI collaboration—when powered by high-precision initial segmentation—can overcome historic barriers in pediatric and 3D dental imaging, accelerating the translation of AI tools into clinical practice. Our challenge catalyzed algorithms that transform dental image annotation from a hours-long manual task to a minutes-long collaborative effort, with profound implications for clinical scalability and global dental health equity. Future work should focus on real-time interactive refinement tools to further harness human-AI synergies. The results underscore the critical role of semi-supervised algorithms in enabling scalable, cost-effective dental diagnostics while maintaining clinical accuracy—a prerequisite for widespread adoption in resource-constrained settings. Also, this study directly addresses two critical gaps in dental AI research: (1) the absence of benchmarks for human-AI collaboration efficiency in pediatric dental imaging, and (2) the lack of comparative data on how semi-supervised model quality impacts real-world clinical labor. By simulating real-world refinement scenarios, we quantify how algorithmic performance translates to tangible clinical time savings and evaluate the viability of semi-supervised solutions for underserved tasks (e.g., pediatric X-rays and CBCT segmentation, where public datasets are absent).

\subsection{Limitations}
While the STS 2024 Challenge made strides in advancing dental image segmentation, it has limitations. The dataset, although diverse, remains limited in size and scope, potentially affecting the generalizability of the findings. Additionally, the challenge primarily focused on segmentation accuracy, with less emphasis on other critical factors such as processing speed and integration with clinical workflows. Furthermore, while representative of real-world scenarios, the high proportion of unannotated data may have introduced biases that could influence the performance of semi-supervised algorithms. The STS 2024 Challenge has been instrumental in highlighting the potential of semi-supervised semantic segmentation in dental medical imaging. By addressing key challenges related to data scarcity and segmentation precision and by introducing innovative evaluation metrics, the challenge has paved the way for future advancements in automated dental analysis. Moving forward, continued efforts to expand and diversify dental datasets and develop more integrated and sophisticated learning algorithms will be essential in realizing the full potential of AI-driven dental diagnostics and treatment planning.

\section{Conclusion}

The STS 2024 Challenge represents an advancement in semi-supervised instance segmentation for dental imaging, establishing the first standardized benchmark for label-efficient learning in this clinically critical domain. Our comprehensive analysis reveals that successful methodologies consistently leveraged synergistic combinations of three core strategies: knowledge transfer from foundation models, iterative pseudo-label refinement, consistency regularization leanring, and multi-stage architectural optimization. The dominance of teacher-student frameworks employed by 85\% of top-ranked solutions underscores their effectiveness in propagating knowledge from limited labeled data through carefully designed pseudo-labeling cycles. Notably, the integration of Segment Anything Model (SAM) variants and nnU-Net backbones emerged as particularly potent, enabling teams like Guo777 and Isjinhao to achieve instance-level Dice scores exceeding 90\% in CBCT segmentation while utilizing merely 9\% labeled data. These technical innovations directly translate to tangible clinical value, as evidenced by our human-machine collaboration study demonstrating that top solutions reduced annotation time by 88.2\% for adult OPG and 94.1\% for CBCT volumes when compared to manual annotation with junior clinicians achieving 73\% of senior-expert efficiency when refining high-quality predictions.

Despite these advances, several critical challenges persist. The underutilization of unlabeled data by approximately 60\% of participating teams highlights a persistent gap between semi-supervised learning potential and practical implementation. Furthermore, while multi-stage pipelines like Haoyuuu's T3Net showed exceptional performance in 3D segmentation, their computational complexity presents deployment barriers in time-sensitive clinical settings. Limitations in dataset diversity, particularly the scarcity of pediatric CBCT scans and complex pathological cases, also constrain model generalizability. Looking forward, three priority directions emerge: 1) Developing hybrid SSL frameworks that combine SAM's zero-shot generalization with uncertainty-aware consistency constraints, 2) Creating cross-modal learning architectures that leverage complementary information from both 2D and 3D dental imaging, and 3) Establishing standardized clinical validation protocols that assess both segmentation accuracy and real-world workflow integration. The publicly released STS 2024 dataset and open-sourced methodologies provide an essential foundation for addressing these priorities. By catalyzing research in label-efficient learning for dental instance segmentation, this initiative accelerates progress toward accessible, AI-enhanced diagnostics that can transform global oral healthcare, particularly in resource-constrained settings where expert annotation remains scarce.

\section*{CRediT authorship contribution statement}


\textbf{Yaqi Wang}: Conceptualization, Investigation, Funding acquisition 
\textbf{Jun Liu}: Supervision, Project administration, Conceptualization, Funding acquisition, Writing – Review \& Editing 
\textbf{Yifan Zhang}: Data Curation, Resources 
\textbf{Shuai Wang}: Supervision, Methodology, Software, Validation, Writing – Review \& Editing 
\textbf{Huiyu Zhou}: Supervision, Project administration, Conceptualization, Writing – Review \& Editing
\textbf{Zhi Li}: Writing – Review \& Editing, Validation, Visualization 
\textbf{Chengyu Wu}: Writing – Original Draft, Validation 
\textbf{Jialuo Chen}: Writing – Review, Validation 
\textbf{Jiaxue Ni}: Investigation, Validation 
\textbf{Qian Luo}: Writing – Review, Validation 
\textbf{Jin Liu}: Investigation, Formal analysis 
\textbf{Can Han}: Investigation, Validation
\textbf{Changkai Ji}: Competitor 
\textbf{Zhi Qin Tan}: Competitor 
\textbf{Ajo Babu George}: Competitor 
\textbf{Liangyu Chen}: Competitor
\textbf{Qianni Zhang}: Investigation, Formal analysis 
\textbf{Dahong Qian}: Investigation, Validation

\section*{Declaration of competing interest}
The authors declare that they have no known competing financial interests or personal relationships that could have appeared to influence the work reported in this paper.

\section*{Acknowledgements}
This work was supported by the National Natural Science Foundation of China (No. 62206242, No. 62201323), Zhejiang Provincial Natural Science Foundation of China (No. LD25F020005), and China Science and Technology Foundation of Sichuan Province (No. 2022YFS0116). There are no conflicts of interest between authors. Yifan Zhang is the principal sponsor of the challenge by collecting and providing clinical data. Only the organizers and members of their immediate team have access to test case labels. The study protocol was approved by the Medical Ethics Committee of Hangzhou Stomatological Hospital (Approval No: 2022YR014).

\section*{Data availability}
Data will be made available on request.

\bibliographystyle{elsarticle-num} 
\bibliography{refs} 

@inproceedings{bolelli2025segmenting,
  title={Segmenting Maxillofacial Structures in CBCT Volumes},
  author={Bolelli, Federico and Marchesini, Kevin and van Nistelrooij, Niels and Lumetti, Luca and Pipoli, Vittorio and Ficarra, Elisa and Vinayahalingam, Shankeeth and Grana, Costantino},
  booktitle={Proceedings of the Computer Vision and Pattern Recognition Conference},
  pages={5238--5248},
  year={2025}
}

@article{bolelli2024segmenting,
  title={Segmenting the inferior alveolar canal in cbcts volumes: the toothfairy challenge},
  author={Bolelli, Federico and Lumetti, Luca and Vinayahalingam, Shankeeth and Di Bartolomeo, Mattia and Pellacani, Arrigo and Marchesini, Kevin and Van Nistelrooij, Niels and Van Lierop, Pieter and Xi, Tong and Liu, Yusheng and others},
  journal={IEEE Transactions on Medical Imaging},
  year={2024},
  publisher={IEEE}
}

@article{ben2022teeth3ds,
  title={Teeth3DS: a benchmark for teeth segmentation and labeling from intra-oral 3D scans},
  author={Ben-Hamadou, Achraf and others},
  journal={arXiv preprint arXiv:2210.06094},
  year={2022}
}

@article{li2024fine,
  title={A fine-grained orthodontics segmentation model for 3D intraoral scan data},
  author={Li, Juncheng and others},
  journal={Computers in Biology and Medicine},
  volume={168},
  year={2024}
}

@inproceedings{zanjani2019deep,
  title={Deep learning approach to semantic segmentation in 3D point cloud intra-oral scans of teeth},
  author={Zanjani, Farhad Ghazvinian and Moin, David Anssari and Verheij, Bas and Claessen, Frank and Cherici, Teo and Tan, Tao and others},
  booktitle={International Conference on Medical Imaging with Deep Learning},
  pages={557--571},
  year={2019},
  organization={PMLR}
}

@article{wang2025miccai,
  title={MICCAI 2023 STS Challenge: A retrospective study of semi-supervised approaches for teeth segmentation},
  author={Wang, Yaqi and Zhang, Yifan and Chen, Xiaodiao and Wang, Shuai and Qian, Dahong and Ye, Fan and Xu, Feng and Zhang, Hongyuan and Dan, Ruilong and Zhang, Qianni and others},
  journal={Pattern Recognition},
  pages={112049},
  year={2025},
  publisher={Elsevier}
}

@inproceedings{cui2022ctooth+,
  title={Ctooth+: A large-scale dental cone beam computed tomography dataset and benchmark for tooth volume segmentation},
  author={Cui, Weiwei and Wang, Yaqi and Li, Yilong and Song, Dan and Zuo, Xingyong and Wang, Jiaojiao and Zhang, Yifan and Zhou, Huiyu and Chong, Bung san and Zeng, Liaoyuan and others},
  booktitle={MICCAI Workshop on Data Augmentation, Labelling, and Imperfections},
  pages={64--73},
  year={2022},
  organization={Springer}
}

@article{huang2024multimodal,
  title={A multimodal dental dataset facilitating machine learning research and clinic services},
  author={Huang, Yunyou and Liu, Wenjing and Yao, Caiqin and Miao, Xiuxia and Guan, Xianglong and Lu, Xiangjiang and Liang, Xiaoshuang and Ma, Li and Tang, Suqin and Zhang, Zhifei and others},
  journal={Scientific Data},
  volume={11},
  number={1},
  pages={1291},
  year={2024},
  publisher={Nature Publishing Group UK London}
}

@ARTICLE{9686728,
  author={Cipriano, Marco and Allegretti, Stefano and Bolelli, Federico and Di Bartolomeo, Mattia and Pollastri, Federico and Pellacani, Arrigo and Minafra, Paolo and Anesi, Alexandre and Grana, Costantino},
  journal={IEEE Access}, 
  title={Deep Segmentation of the Mandibular Canal: A New 3D Annotated Dataset of CBCT Volumes}, 
  year={2022},
  volume={10},
  number={},
  pages={11500-11510},
  doi={10.1109/ACCESS.2022.3144840}}

@ARTICLE{9083982,
  author={Chen, Yanlin and Du, Haiyan and Yun, Zhaoqiang and Yang, Shuo and Dai, Zhenhui and Zhong, Liming and Feng, Qianjin and Yang, Wei},
  journal={IEEE Access}, 
  title={Automatic Segmentation of Individual Tooth in Dental CBCT Images From Tooth Surface Map by a Multi-Task FCN}, 
  year={2020},
  volume={8},
  number={},
  pages={97296-97309},
  doi={10.1109/ACCESS.2020.2991799}}

@article{isensee2021nnu,
  title={nnU-Net: a self-configuring method for deep learning-based biomedical image segmentation},
  author={Isensee, Fabian and Jaeger, Paul F and Kohl, Simon AA and Petersen, Jens and Maier-Hein, Klaus H},
  journal={Nature methods},
  volume={18},
  number={2},
  pages={203--211},
  year={2021},
  publisher={Nature Publishing Group}
}

@article{Elgarba2024_AI_CBTC_IOS_registration,
  title = {Validation of a novel AI-based automated multimodal image registration of CBCT and intraoral scan aiding presurgical implant planning},
  author = {Elgarba, B. M. and Fontenele, R. C. and Ali, S. and Swaity, A. and Meeus, J. and Shujaat, S. and Jacobs, R.},
  journal = {Clinical Oral Implants Research},
  year = {2024},
  doi = {10.1111/clr.14338},
  pmid = {39101603}
}

@article{Kim2023_AutoReg_CBTC_IOS_segmentation,
  title = {Novel Procedure for Automatic Registration between Cone-Beam Computed Tomography and Intraoral Scan Data Supported with 3D Segmentation},
  author = {Kim, Yoon-Ji and Ahn, Jang-Hoon and Lim, Hyun-Kyo and Nguyen, Thong Phi and Jha, Nayansi and Kim, Ami and Yoon, Jonghun},
  journal = {Bioengineering},
  volume = {10},
  number = {11},
  pages = {1326},
  year = {2023},
  doi = {10.3390/bioengineering10111326}
}

@article{Wang2023_RootCanalSegmentation_DentalNet_PulpNet,
  title = {Root canal treatment planning by automatic tooth and root canal segmentation in dental CBCT with deep multi-task feature learning},
  author = {Wang, Yiwei and Xia, Wenjun and Yan, Zhennan and Liu, Dong and Zhang, Tianqi and Chen, Zhen and Wei, Sheng and Zhang, Yuxuan},
  journal = {Medical Image Analysis},
  year = {2023},
  pmid = {36682153},
  note = {Elsevier}
}

@inproceedings{3DUNet_GlobalLocal_Loss_RootCanal2021,
  title = {Root Canal Segmentation in CBCT Images by 3D U-Net with Global and Local Combination Loss},
  author = {Zhang, Jian and Xia, Wenjun and Dong, Jiaqi and Tang, Zisheng and Zhao, Qunfei},
  booktitle = {Proceedings of the IEEE Engineering in Medicine and Biology Society (EMBC)},
  year = {2021},
  pmid = {34891897}
}

@article{RefinedPulp_UNet_2021,
  title = {Refined tooth and pulp segmentation using U-Net in CBCT image},
  author = {Duan, Wei and Chen, Yufei and Zhang, Qi and Lin, Xiang and Yang, Xiaoyu},
  journal = {Dentomaxillofacial Radiology},  
  year = {2021},
  pmid = {33444070}
}

@article{Schulze2016_CBCT_IOS_registration,
  title = {Registration of cone beam computed tomography data and intraoral surface scans: A prerequisite for guided implant surgery with CAD/CAM drilling guides},
  author = {Fl{\"u}gge, Tabea and Derksen, Wiebe and Te Poel, Jobine and Hassan, Bassam and Nelson, Katja and Wismeijer, Daniel},
  journal = {Clinical Oral Implants Research},
  volume = {23},
  pages = {416--423},
  year = {2016},
  doi = {10.1111/clr.12925},
  pmid = {27440381}
}

@article{SystematicReview_CBCT_IOS_registration_2025,
  title = {Automatic multimodal registration of cone-beam computed tomography and intraoral scans: a systematic review and meta-analysis},
  author = {Zheng, Qianhan and Wu, Yongjia and Chen, Jiahao and Wang, Xiaozhe and Zhou, Mengqi and Li, Huimin and Lin, Jiaqi and Zhang, Weifang and Chen, Xuepeng},
  journal = {Clinical Oral Investigations},
  year = {2025},
  pmid = {39878846}
}

\end{document}